\documentclass[lettersize,journal]{IEEEtran}
\usepackage{amsmath,amsfonts}
\usepackage{algorithmic}
\usepackage{array}
\usepackage[caption=false,font=normalsize,labelfont=sf,textfont=sf]{subfig}
\usepackage{textcomp}
\usepackage{stfloats}
\usepackage{url}
\usepackage{verbatim}
\usepackage{graphicx}
\usepackage{hyperref}   
\usepackage{cleveref}
\usepackage{colortbl}  
\usepackage[dvipsnames]{xcolor} 
\definecolor{c1}{HTML}{81B29A}	
\definecolor{c2}{HTML}{3D405B}
\usepackage{xcolor}
\usepackage{makecell}
\usepackage{array}   
\usepackage{arydshln} 
\usepackage{amsmath}
\usepackage{multirow}   
\usepackage{graphicx}
\usepackage{booktabs}
\usepackage{soul}   
\usepackage{graphicx}
\usepackage{csquotes} 
\def\BibTeX{{\rm B\kern-.05em{\sc i\kern-.025em b}\kern-.08em
    T\kern-.1667em\lower.7ex\hbox{E}\kern-.125emX}}
\usepackage{balance}
\begin{document}
	
\title{{AI-T2I: Aggregating-and-Isolating Cross-Attention to Diffusion Models for Text-to-Image Synthesis}}
\author{Shipeng Cao, Biao Qian$^{*}$, Haipeng Liu, Yang Wang$^{*}$, \textit{Senior Member}, \textit{IEEE}, Meng Wang, \textit{Fellow}, \textit{IEEE}
	\thanks{This research is supported by Institute of Advanced Medicine and Frontier Technology (2023IHM01080), and sponsored by CCF-NetEase ThunderFire Innovation Research Funding (NO. CCF-Netease 202513); The computation is completed on the HPC Platform of Hefei University of Technology.}
	
	\thanks{Shipeng Cao, Haipeng Liu, Yang Wang and Meng Wang are with Key Laboratory of Knowledge Engineering With Big Data, Ministry of Education, Hefei University of Technology, China. E-mail: caoshipeng@mail.hfut.edu.cn, hpliu\_hfut@hotmail.com, yangwang@hfut.edu.cn, eric.mengwang@gmail.com.}
	\thanks{Biao Qian is with Department of Automation, Tsinghua University, China. E-mail: hfutqian@gmail.com.}
	
	\thanks{Corresponding authors ($^{*}$): Biao Qian, Yang Wang}	
}


\markboth{IEEE TRANSACTIONS ON MULTIMEDIA}%
{Cao \MakeLowercase{\textit{et al.}}: Aggregating-and-Isolating Cross-Attention to Diffusion Models for Text-to-Image Synthesis}

\maketitle


\begin{abstract}
	Text-to-image synthesis has made significant progress, benefiting from the strong generative capabilities of diffusion models. However, these models struggle to achieve precise text-to-image alignment within cross-attention maps during the denoising process. Existing works primarily focus on inter-subject-token activations (\emph{i.e.}, cross-attention scores) overlap for different subjects, overlooking the intra-subject-token activations scattering issue for identical subjects. In this paper, we propose an \underline{A}ggregating-and-\underline{I}solating cross-attention approach to diffusion models for \underline{T}ext-to-\underline{I}mage {s}ynthesis, dubbed AI-T2I. Technically, to address the scattering issue, we devise an aggregation loss to identify and consolidate the scattered intra-token activations, which implicitly helps mitigate the potential overlap issue. Upon that, an isolation loss is further introduced to push the inter-token activations apart, thus fulfilling precise text-to-image alignment.  Extensive experiments on various benchmarks demonstrate the superiority of AI-T2I over the state-of-the-art works for text-to-image synthesis. Furthermore, our AI-T2I exhibits excellent generalization across other tasks, \emph{e.g.}, controllable layout generation and personalized generation. \emph{Our code is available at https://github.com/Hatter77/AI-T2I}.
\end{abstract}


\section{Introduction}
\label{sec:intro}
\begin{figure}[t]
	\centering
	\setlength{\abovecaptionskip}{-0.1cm}
	\setlength{\belowcaptionskip}{-0.4cm}
	\includegraphics[width=0.8\linewidth]{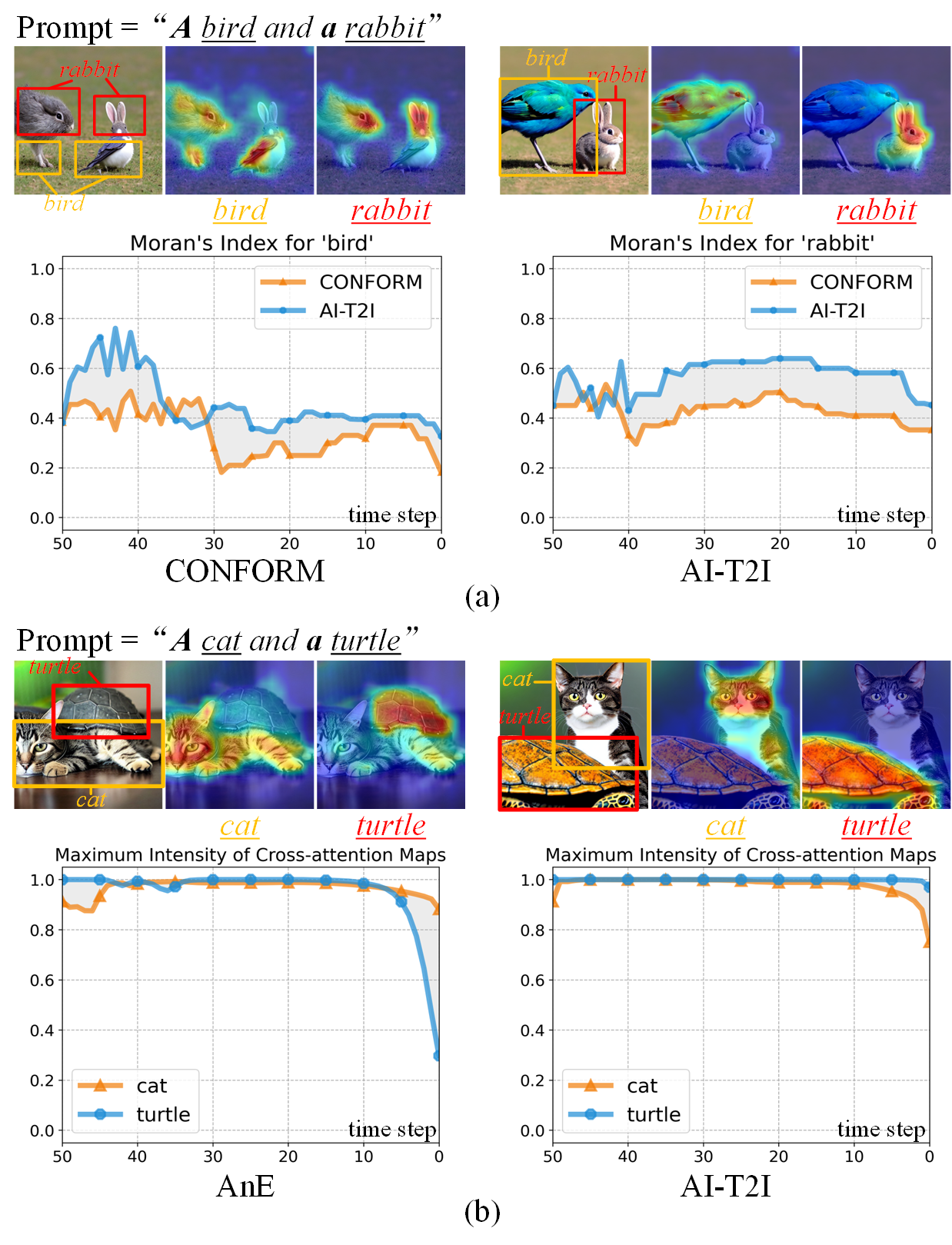}
	\caption{Existing methods, \emph{e.g.}, (a) CONFORM ~\cite{conform} fails to tackle the intra-token scattering due to the reliance on the appropriate initial noise, resulting in multiple high-intensity activation regions within the cross-attention map corresponding to identical subjects. Moran's Index quantified attention map scattering, with values closer to 1 indicating less scattering; (b) AnE ~\cite{attend} enhances the appearance of subjects by maximizing activation intensities; however, it is less effective at subject mixing. In contrast, our AI-T2I simultaneously addresses intra-token activations scattering and inter-token overlap, yielding desirable subject generation results.}
	
	\label{fig:other_fails}
\end{figure}

\IEEEPARstart{T}{ext-to-image (T2I)} synthesis has attracted considerable attention and significant efforts have been made to resolve the issue. Among the existing generative models applied in various fields \cite{Qian_2023_CVPR,qian2023rethinking,10476709,qian2022switchable}, diffusion models \cite{song2020denoising,dhariwal2021diffusion,dong2023prompt,mao2023guided,Liu_2024_CVPR,One_Stone,liu2026chipdiff,zhang2025survey} have been widely adopted to benefit from the powerful denoising process. Specifically, given the pure initial noise, the denoising process is performed by yielding cross-attention maps for text-to-image alignment. However, the activations (i.e. cross-attention scores) within the cross-attention map may fail to exhibit the ideal distribution based on prompt tokens, such as intra-subject-token$\footnote{Tokens mainly include subject (e.g., animals) and attribute (e.g., colors) types. As indicated in \cite{attend,zhang2024object}, the alignment of attribute tokens strongly depends on subject tokens. Hence, we mainly consider the subject token, and abbreviate intra-subject-token as intra-token for the rest of the paper.}$ activations scattering \cite{conform, astar}, where the attention activations belonging to one subject token are spatially fragmented across the attention map; thereby parts of the same subject are generated in spatially disconnected locations, producing fragmented or duplicated object instances. This issue may lead to parts of a single object to become scattered across different locations. For example, as shown in Fig. \ref{fig:other_fails}(a), the ears of the rabbit are scattered onto the body of the bird instead of being correctly formed with the body of the rabbit, contrary to the prompt “a bird and a rabbit”. To resolve this issue, A-star \cite{astar} and CONFORM \cite{conform} attempt to address intra-token scattering; however, these models heavily rely on the initial noise. During each denoising step, they aim to maintain consistency between the current cross-attention map and that of the previous time step.

Due to the random nature of initial noise, the high-intensity activations within the cross-attention map varied a lot during each step, hence may easily result in activations scattering for intra-tokens. Furthermore, we hypothesize that intra-token scattering acts as one of the primary drivers behind inter-token overlap\footnote{Akin to intra-token, inter-token in this paper refers to the subject rather than the attribute within a text prompt.}, where attention activations for different subject tokens occupy overlapping spatial regions, leading to subjects mixing or sharing attributes. To resolve inter-token overlap, current approaches \cite{attend,astar,li2023divide,zhang2024object,linguistic,conform} predominantly exploit cross-attention mechanisms that align textual tokens with visual content during image synthesis. Specifically, Attend-and-Excite (AnE) \cite{attend} enhances the appearance of subjects in generated images by maximizing activation intensities; however, it is less effective at mitigating mixing. For instance, as seen in Fig. \ref{fig:other_fails}(b), a cat blends with a turtle rather than being given the prompt “a cat and a turtle”. Subsequent studies \cite{astar,li2023divide,zhang2024object,linguistic} address this limitation by optimizing inter-token activations distribution divergence through cosine similarity or KL divergence to reduce inter-token activations overlaps. But prior methods predominantly focus on inter-token overlap while overlooking intra-token scattering. Meanwhile, previous methods \cite{conform,zhang2024object} relying on cosine similarity, are built upon the CLIP model \cite{clip} along with the Stable Diffusion (SD) models \cite{sd} over latent representations.

To address the above issues, in this paper, we propose AI-T2I, a novel method to aggregate and isolate cross-attention activations in diffusion models for text-to-image synthesis. Our basic idea is to tackle intra-token scattering and inter-token
overlap issues for precise text-to-image alignment, which is achieved through dual complementary objectives. Since both issues stem from incoherent spatial distributions of activations in cross-attention maps, our key insight is to form separate, well-defined high-intensity activation regions within cross-attention maps for each subject during the denoising process. To this end, an aggregation loss is first devised to consolidate the scattered intra-token activations from identical subjects, which, meantime, is beneficial for mitigating the potential inter-token overlap. Building on that, the isolation loss is utilized to further isolate the inter-token activations from different subjects. As a byproduct, we perform further analysis on the distance metrics, \emph{i.e.}, cosine similarity and Euclidean distance, to strengthen the aggregation and isolation losses. Empirically, AI-T2I is evaluated on standard benchmarks, \emph{i.e.}, the AnE \cite{attend} and T2I-CompBench++ \cite{t2ibench} datasets, demonstrating consistent improvements over state-of-the-art works. Additionally, we exhibit the plug-and-play versatility of our AI-T2I by integrating it into other diffusion-based frameworks across downstream tasks, such as controllable layout generation and personalized synthesis.

Our core contributions are summarized as follows: (1) We propose a novel AI-T2I, to tackle both intra-token scattering and inter-token overlap issues in cross-attention maps through the proposed aggregation and isolation losses. (2) We provide an in-depth analysis of distance metrics (e.g., cosine similarity, Euclidean distance) to enhance aggregation and isolation losses. (3) We empirically demonstrate the superiority and plug-and-play versatility of our AI-T2I by achieving state-of-the-art results on standard benchmarks and showcasing its effectiveness across various downstream tasks.

\section{Related Work}
\label{sec:relate}
Generative adversarial networks (GANs) \cite{hou2025clip,cheng2022vision,yang2024dmf,jiang2025cmsl}, variational autoencoders (VAEs) \cite{huang2017real} and autoregressive models \cite{ding2021cogview,ramesh2021zero} have been adopted for text-to-image synthesis. However, diffusion models \cite{dhariwal2021diffusion,guo2024initno,predicated,linguistic} have recently emerged as the dominant approach for both conditional and unconditional image. In conditional synthesis, classifier-free guidance \cite{dhariwal2021diffusion} improves the fidelity of generated images with respect to the input prompts.  Building on this foundation, subsequent efforts \cite{li2023divide,zhang2025enhancing,binyamin2024make,dahary2024yourself,zhang2025aligning} have further enhanced the consistency between textual conditions and the synthesized images. Some studies attribute misalignment issues to the difficulty of model in understanding complex logical relationships in long sentences. \cite{linguistic} partitions prompt tokens into positive and negative pairs based on part-of-speech, optimizing cross-attention maps to align with syntactic relationships. However, this method degrades to standard SD when prompts lack adjective-noun pairs. \cite{structure} improves text compositionality by splitting long sentences into multiple shorter ones and merging their corresponding synthesis outputs into one image. However, this method requires additional sentence division and cross-attention map fusion. \cite{predicated} represents the intended meanings in a text prompt using predicate logic, offering guidance for text-based image generation. However, excessive constraints on each token pair may lead to suboptimal performance. In contrast, our method applies only a single constraint per token pair, mitigating this issue.

Beyond text-to-image synthesis, a wide array of related tasks have been explored. For instance, some studies \cite{dahary2024yourself,zheng2025semantic,taghipour2025box} employ bounding boxes as auxiliary conditions to guide image layout, while image editing approaches \cite{orgad2023editing,huang2024smartedit,nam2024dreammatcher,guo2024smooth,liu2024towards} modify existing images by adjusting textual constraints to selectively alter content. For example, \cite{zheng2025semantic} regulates object layout in synthesized images by reinforcing attention scores inside the bounding box while attenuating those outside. Furthermore, \cite{xu2024sgdm} extracts features from the input style image, which are then injected into the noise generation process of a Stable Diffusion (SD) model to achieve personalized styling. Most of the above research attempts to address different image generation tasks. In particular, our method further exhibits excellent generalization on various generation tasks. 


\section{Preliminary}
\subsection{Latent Diffusion Models}
We apply our method to the open-source Stable Diffusion (SD) (\cite{sd}), which generates high-quality images from noise. SD operates in three key stages. First, an auto-encoder is trained to map an image $x$ into a latent representation $z = {\cal E}(x)$, where
${\cal E}$ denotes the encoder. Second, the diffusion process is applied to $z$. The forward diffusion process gradually adds Gaussian noise, transforming $z$ into a noisy latent $z_t$, where the time step 
$t$ is uniformly sampled from $\{ 1, \cdots ,T\} $. A neural network then learns iteratively to denoise $z_t$ in a new sample $z_0$; this denoising stage is the reverse process. The conditional reverse process is defined as
\begin{equation}
	\mathbb{E}_{\mathcal{E}(x), y, \epsilon \sim N(0,1), t} \left[ ||\epsilon - \epsilon_{\theta}(z_t, t, \tau_{\theta}(y))||_2^2 \right],
\end{equation}
where $\epsilon$ is Gaussian noise sampled from a standard normal distribution; $\epsilon_{\theta}$ is a parameterized denoising network; $\tau_{\theta}$ is a frozen text encoder and $y$ represents text prompts. During sampling, $z_T$ is randomly sampled from a standard Gaussian distribution and iteratively denoised by $\epsilon_{\theta}$ to produce $z_0$. Finally, a decoder ${\cal D}$ reconstructs $z_0$ into an image $\hat x$ of a specific size, mathematically expressed as $\hat x = {\cal D}(z_0)$.
\subsection{Cross-Attention Mechanism}
SD introduces the cross-attention mechanism \cite{wu2018and,liu2025few,wang2022progressive} to guide image synthesis using text prompts. The cross-attention map $A$ is defined as
\begin{equation} 
	A=\text{softmax}\left(\frac{\mathbf{Q}\mathbf{K}^T}{\sqrt{d_k}}\right),
	\label{eq:eq2}
\end{equation}
where \( d_k \) is the dimensionality of the keys. In SD, $Q$ (query) is mapped from the intermediate image representation, while $K$ (key) is linearly projected from the conditioning text embedding. A pretrained CLIP text encoder is used to encode the text prompt into a sequential text embedding.

\begin{figure*}[t]
	\centering
	\setlength{\abovecaptionskip}{-0.05cm}
	\setlength{\belowcaptionskip}{-0.4cm}
	\includegraphics[width=0.95\linewidth]
	{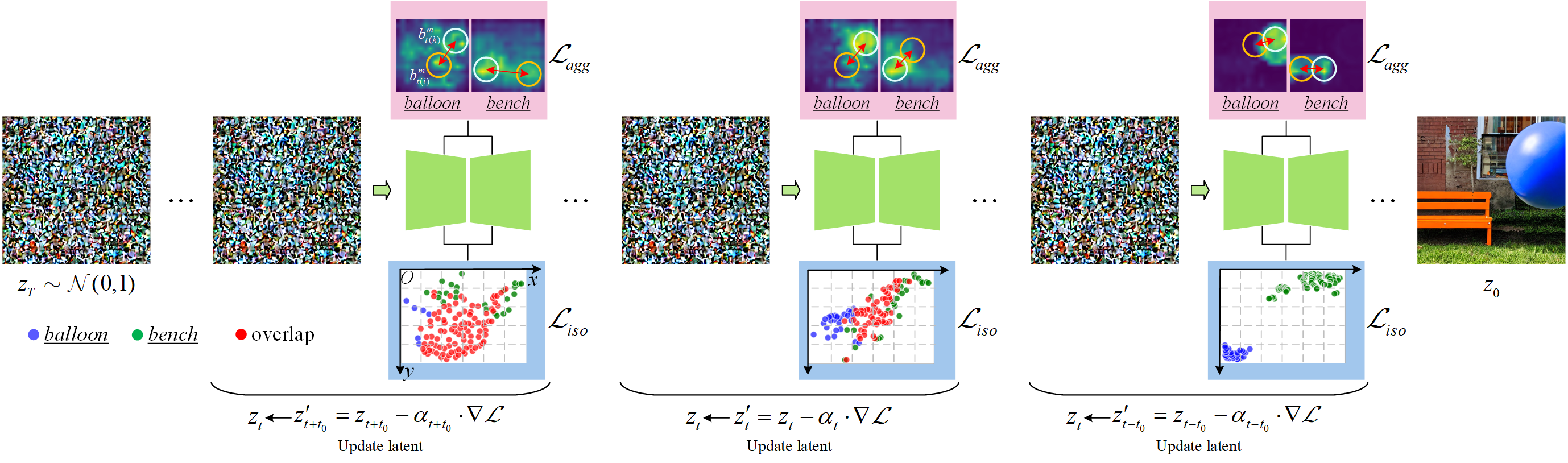}
	\caption{Illustration of our proposed AI-T2I. During the denoising process, the latent $z_t$ at $t$-th timestep is updated to achieve precise text-to-image alignment via aggregation loss (${\cal L}_{agg-sub}+{\cal L}_{agg-attr}$) to address intra-token scattering and isolation loss (${\cal L}_{iso}$) to reduce inter-token overlap.}
	\label{fig:pipeline}
\end{figure*}

\section{AI-T2I: Aggregating-and-Isolating Intra-token and Inter-token Activations}
\label{sec:method}

As discussed in Sec. \ref{sec:intro}, existing methods dominantly address inter-token activations overlap via the loss that maximizes the distances over inter-token activations, measured by similarity metric, 
while overlooking intra-token scattering. To address the above issues, we propose to aggregate-and-isolate intra-token and inter-token activations over cross-attention maps for text-to-image synthesis, named AI-T2I. 
Specifically, we elaborate our proposed AI-T2I from several aspects, including: an aggregation loss to address intra-token activations scattering (Sec. \ref{sec:agg_loss}), an isolation loss to mitigate inter-token activations overlap  (Sec. \ref{sec:isolation_loss}), together with the overall optimization objective  (Sec. \ref{optimization_obj}).
In particular, we operate our method on 16 × 16 cross-attention maps, as they have been shown to contain the most semantic information \cite{prompt2prompt}. Each map contains 256 activations with each representing a cross-attention score calculated by Eq. \ref{eq:eq2}. Each activation has a spatial location within cross-attention map, represented as a two-dimensional coordinate in the cross-attention score matrix. It is noteworthy that each token corresponds to one cross-attention map with the same size during denoising process, we apply both aggregation and isolation losses for intra-token and inter-token activations to directly adopt their value and spatial locations as per the cross-attention map they belong to.

\subsection{Aggregation Loss: Addressing Intra-token Activations Scattering}
\label{sec:agg_loss}

To aggregate activations, we first need to represent the spatial location of each activation. Then, we reduce their spatial distance to address intra-token scattering. Typically, higher-intensity activations for intra-token are considered more representative. Therefore, we aggregate them toward the activation with the maximum value. In the following, we provide a detailed discussion.

\subsubsection{Identifying intra-token activations} As AnE\cite{attend} illustrates, the highest-intensity activation best represents the subject, and high-intensity activations are generally locate around it. Thus, we identify high-intensity activations start from the highest one. Specifically, given a subject token $m$, $A_t^m$ denotes its corresponding cross-attention map at the $t$-th timestep. A global search over $A_t^m$ is performed to identify the highest-intensity activation, represented by ${\max (A_t^m)}$. Then, centered at ${\max (A_t^m)}$, a grouping region is formed to gather a series of activations using a circular mask with radius $r$. Similarly, outside this initial grouping region, a series of grouping regions with radius $r$ are sequentially constructed using masks $\{M^m_{t(i)}\}_{i=1}^{N^c}$, expressed as $\{a^m_{t(i)}\}_{i=1}^{N^c}$, where $a^m_{t(i)}=A_t^m\cdot M^m_{t(i)}$ and $N^{c}$ is the number of grouping regions. The grouping region $a^m_{t(i)}$ comprises activations designated as $a^m_{t(i)} =\{a^m_{t(i,j)}\}_{j=1}^{N_{t(i)}^a}$, with $N_{t(i)}^a$ indicates the number of activations in the $i$-th grouping region at the $t$-th timestep.

\subsubsection{Aggregating identified intra-token activations} 
To address intra-token activations scattering, we propose an aggregation loss  to aggregate varied circular regions with representative high-intensity activations. Specifically, for the activations $a^m_{t(i)}$ from $i$-th circular region, we first calculate their center $b_{t(i)}^m=centroid(a^m_{t(i)})=(h_{t(i)}^m,w_{t(i)}^m)$, where $h_{t(i)}^m$, $w_{t(i)}^m$ serve as two dimensional spatial coordinates over cross-attention map for subject $m$, which are formulated as
\begin{equation} \label{eq3}
	\begin{aligned}
		\begin{split}
			&h_{t(i)}^m=\sum\limits_{a_{t(i,j)}^m \in a_{t(i)}^m} {h_{t(i,j)}^m \cdot {\frac{{ v(a_{t(i,j)}^m)}}{{\sum\limits_{a_{t(i,j)}^m \in a_{t(i)}^m} {v(a_{t(i,j)}^m)} }}} }, \\
			&w_{t(i)}^m=\sum\limits_{a_{t(i,j)}^m \in a_{t(i)}^m} {w_{t(i,j)}^m \cdot  {\frac{{v(a_{t(i,j)}^m)}}{{\sum\limits_{a_{t(i,j)}^m \in a_{t(i)}^m} {v(a_{t(i,j)}^m)} }}} },
		\end{split}
	\end{aligned}
\end{equation}
where ($h_{t(i,j)}^m$, $w_{t(i,j)}^m$) denotes the two dimensional spatial coordinates over cross-attention map for $j$-th activation $a_{t(i,j)}^m$ from $i$-th circular region, and $v(\cdot)$ represents the activation value function.
With the centers $b_{t(i)}^m, i \in \{1,2,...,N^{c}\}$ from all circular regions, we formulate the aggregation loss for subject $m$ within the $t$-th step as
\begin{equation}
	\begin{aligned}
		{{\cal L}_{agg - sub}} &= \sum\limits_{i,k = 1,i \ne k}^{{N^c}} {Dist(b_{t(i)}^m,b_{t(k)}^m)} \\ &= \sum\limits_{i,k = 1,i \ne k}^{{N^c}} {\sqrt {{{(h_{t(i)}^m - h_{t(k)}^m)}^2} + {{(w_{t(i)}^m - w_{t(k)}^m)}^2}} }
	\end{aligned}
	\label{eq:agg_sub}
\end{equation}
where $Dist(\cdot)$ stands for the distance metric function; we adopt the Euclidean function in our experiments, which is further discussed in Sec.\ref{sec:fur_inves}.

\subsubsection{Aggregation loss for attribute} 
As indicated \cite{attend,zhang2024object}, the attribute token is closely relying on subject token within denoising process for image synthesis. Hence, to synthesize the image upon attribute token, we resort to aggregation loss, denoted as ${\cal L}_{agg-attr}$, rather than isolation loss, owing to the same subject they both correspond to.


\begin{table*}[t]
	\small
	\centering
	\caption{Comparison of Euclidean distance and cosine similarity towards the aggregation and isolation losses. $\uparrow$: Higher is better.  The best and second-best results are reported with \textbf{boldface} and \underline{underline}, respectively.}
	\resizebox{\textwidth}{!}{ 
		\begin{tabular}{l c c c c c c c c c} 
			\toprule
			
			\multirow{2}{*}{\textbf{Case}} & \multicolumn{3}{c}{\textbf{Object-Object}} & \multicolumn{3}{c}{\textbf{Animal-Object}} & \multicolumn{3}{c}{\textbf{Animal-Animal}} \\
			& Full Sim. (\(\uparrow\)) & Min. Sim. (\(\uparrow\)) & T-C Sim. (\(\uparrow\)) & Full Sim. (\(\uparrow\)) & Min. Sim. (\(\uparrow\)) & T-C Sim. (\(\uparrow\)) & Full Sim. (\(\uparrow\)) & Min. Sim. (\(\uparrow\)) & T-C Sim. (\(\uparrow\)) \\
			\midrule
			
			\rowcolor{gray!30}
\multicolumn{10}{@{}l}{\textbf{A}: AI-T2I \emph{w/o} ($\mathcal{L}_{\text{iso}}+\mathcal{L}_{\text{agg-attr}}$)} \\
			Cos    &0.348   &0.260  &0.769  &0.358  &0.267  &0.820  &0.331  &0.244  &0.792 \\
			Euc    &0.364 &0.272  &0.822 &\textbf{0.364}  &\textbf{0.271}  &0.843  &0.334  &0.248  &0.804 \\             
			\midrule
			
			\rowcolor{gray!30}
\multicolumn{10}{@{}l}{\textbf{B}: AI-T2I \emph{w/o} ($\mathcal{L}_{\text{agg-sub}}+\mathcal{L}_{\text{agg-attr}}$)} \\
			Cos    &0.362  &0.268  &0.822  &0.358  &0.267  & 0.848 & 0.338 &0.255  & 0.815 \\
			Euc    &0.362  &0.269  &0.823  &0.359  &0.267  &0.844  & \underline{0.339} &0.255 &0.833 \\                 
			\midrule

			\rowcolor{gray!30}
			\multicolumn{10}{@{}l}{\textbf{C}: AI-T2I \emph{w/o} $\mathcal{L}_{\text{agg-attr}}$} \\
			Cos+Cos & 0.364 & 0.270 & 0.826 & 0.360 & 0.268 & \underline{0.846} & 0.339 & 0.256 & 0.821 \\
			Cos+Euc &0.364   &0.271 &0.819  &0.360  &0.268  &0.842  &\underline{0.340}  &0.255  &\textbf{0.837} 
			
			\\
			Euc+Cos &\underline{0.366}   &\underline{0.273} &\textbf{0.829}  &0.361  &\underline{0.270}  &0.845  &\underline{0.340}  &\textbf{0.257} &0.815 \\
			
			Euc+Euc & \textbf{0.368} & \textbf{0.274} & \underline{0.828} & \underline{0.363} & 0.269 & \textbf{0.847} & \textbf{0.342} & \textbf{0.257} & \underline{0.834} \\
			\bottomrule
		\end{tabular}
	}
	\label{tb:blip_cos_euc_compare_full}
\end{table*}

\subsection{Isolation Loss: Reducing Inter-token Activations overlap}
\label{sec:isolation_loss}
The above aggregation loss is capable of consolidating the scattered intra-token activations to obtain a whole subject, which may potentially mitigate the overlap issue for different subjects.
To further reduce inter-token overlap for precise text-to-image alignment, we introduce an isolating loss to push the inter-token activations apart. Specifically, for subjects $m$ and $n$ in the prompt token set $S$, let 
$A_t^m$ and $A_t^n$ denote their corresponding cross-attention maps at timestep $t$, then the isolation loss is calculated as
\begin{equation} \label{eq:l_ios}
	{{\cal L}_{iso}} = 1 - d/{d_{\max }},
\end{equation}
where $d$ denotes the inter-subject distance, computed as
\begin{equation}
	\begin{aligned}
			d &= Dist(centroid(A_t^m),centroid(A_t^n)) \\&=Dist((h_t^m, w_t^m),(h_t^n, w_t^n)) \\&=
			\sqrt {{{(h_t^m - h_t^n)}^2} + {{(w_t^m - w_t^n)}^2}},
	\end{aligned}
\end{equation}
where $centroid(\cdot)$ returns the centroid coordinates of attention map $A_t^s$ for $s \in {m, n}$ (refer to Eq. \ref{eq3}); $Dist(\cdot)$ stands for the distance metric function. To prevent excessive inter-subject distances from pushing subjects toward the image boundaries, $d$ is normalized by
\begin{equation}
	d_{\max} = \sqrt{W^2 + H^2},
\end{equation}
where $W$ and $H$ are the attention map width and height. The objective of Eq. \ref{eq:l_ios} is to maximize the distance between activations of distinct subjects to alleviate inter-token activations overlap.


\subsection{Optimizing Latent Diffusion Models with Aggregation and Isolation Losses}
\label{optimization_obj}
To support the maximum-value activation as a reference to locate intra-token activations, we adopt the ${{\cal L}_{\max }}$ loss following \cite{attend,zhang2024object,phung2024grounded}, which is designed to strengthen the maximum value in attention maps, formulated as
\begin{equation}
	{{\cal L}_{\max }} = 1 - v(\max (A_t^m))
	\label{l_max}
\end{equation}
where $\max(\cdot)$ is the maximum value function. Together with Eq. \ref{l_max}, the overall loss function contains both attention isolation loss and aggregation loss as
\begin{equation}
	{\cal L} = {{\lambda _1}{{\cal L}_{agg-sub}}+{\lambda _2}{{\cal L}_{iso}} + {\lambda _3}{{\cal L}_{max}}+{\lambda _4}{\cal L}_{agg-attr}},
	\label{eq:overall_loss}
\end{equation}
where $\lambda _1$, $\lambda _2$, $\lambda _3$ and $\lambda _4$ denote the balancing parameters.
We now only need to direct the latent code at the current time step $z_t$ in the right direction as measured by this overall loss (Eq. \ref{eq:overall_loss}). We realize this with a latent update: ${z'_t} = {z_t} - {\alpha _t} \cdot {\nabla _{{z_t}}}{\cal L}$, where $\alpha_t$ 
is the step size of the gradient update. This updated $z'_t$ is then used in the next denoising step to obtain $z_{t-1}$, which is then updated in a similar fashion, and the process repeats until the final step.


\begin{table*}[t]
	\small
	\centering
	\caption{{Comparison of quantitative results (\emph{i.e.}, \emph{Full Sim.}, \emph{Min. Sim.} and \emph{T-C Sim.}) with state-of-the-art works on the AnE dataset.} $\uparrow$: Higher is better. The best and second-best results are reported with \textbf{boldface} and \underline{underline}, respectively.} 
	\resizebox{\textwidth}{!}{ 
		\begin{tabular}{@{}l c c c c c c c c c@{}} 
			\toprule
			\multirow{2}{*}{\textbf{Method}} & \multicolumn{3}{c|}{\textbf{Object-Object}}   & \multicolumn{3}{c|}{\textbf{Animal-Object}} & \multicolumn{3}{c|}{\textbf{Animal-Animal}} \\
			~ & Full Sim. (\(\uparrow\))  & Min. Sim. (\(\uparrow\)) & T-C Sim. (\(\uparrow\)) & Full Sim. & Min. Sim. & T-C Sim. & Full Sim. & Min. Sim. & T-C Sim. \\
			\midrule
			StableDiffusion \cite{sd}  & 0.335 & 0.235 & 0.765 & 0.340 & 0.246 & 0.793 & 0.311 & 0.213 & 0.767  \\
			Structure \cite{structure}  & 0.332 & 0.234 & 0.762 & 0.336 & 0.242 & 0.781 & 0.306 & 0.210 & 0.761  \\
			AnE \cite{attend} & \underline{0.360} & \underline{0.270} & 0.811 & 0.353 & 0.265 & 0.830 & 0.332 & 0.248 & 0.806  \\
			
			SG \cite{linguistic}  & 0.355 & 0.262 & 0.811 & 0.355 & 0.264 & 0.830 & 0.311 & 0.213 & 0.767  \\
			EBCA \cite{park2024energy}  & 0.321 & 0.231 & 0.726 & 0.317 & 0.229 & 0.732 & 0.291 & 0.215 & 0.722  \\
			PredicatedDiff \cite{predicated}  & 0.358 & \underline{0.270} & 0.805 & 0.357 & 0.265 & 0.834 & 0.338 & 0.250 & 0.823  \\
			CONFORM \cite{conform}  & 0.358 & 0.266 & \underline{0.820} & 0.358 & \underline{0.267} & 0.846 & \underline{0.339} & \underline{0.254} & 0.819  \\      
			Atten\&Regulation \cite{zhang2025enhancing} & \underline{0.360} & 0.264 & 0.812 & \underline{0.363} & 0.266 & 0.841 & 0.329 & 0.238 & 0.806  \\ 
			Self-cross \cite{2025selfcross} & 0.359 & 0.267 & 0.825 & 0.351 & 0.261 & \underline{0.847} & 0.332 & 0.251 & \textbf{0.843}  \\ 
			\midrule    
			\rowcolor{c1!40} 
			\textbf{\centering AI-T2I (Ours)} & \textbf{0.369} & \textbf{0.275} & \textbf{0.838} & \textbf{0.364} & \textbf{0.272} & \textbf{0.849} & \textbf{0.342} & \textbf{0.257} & \underline{0.834}  \\
			\bottomrule 
		\end{tabular}
	}
	
	\label{tb:metric_results}
\end{table*}

\begin{figure}[t]
	\centering
	\includegraphics[width=\linewidth]{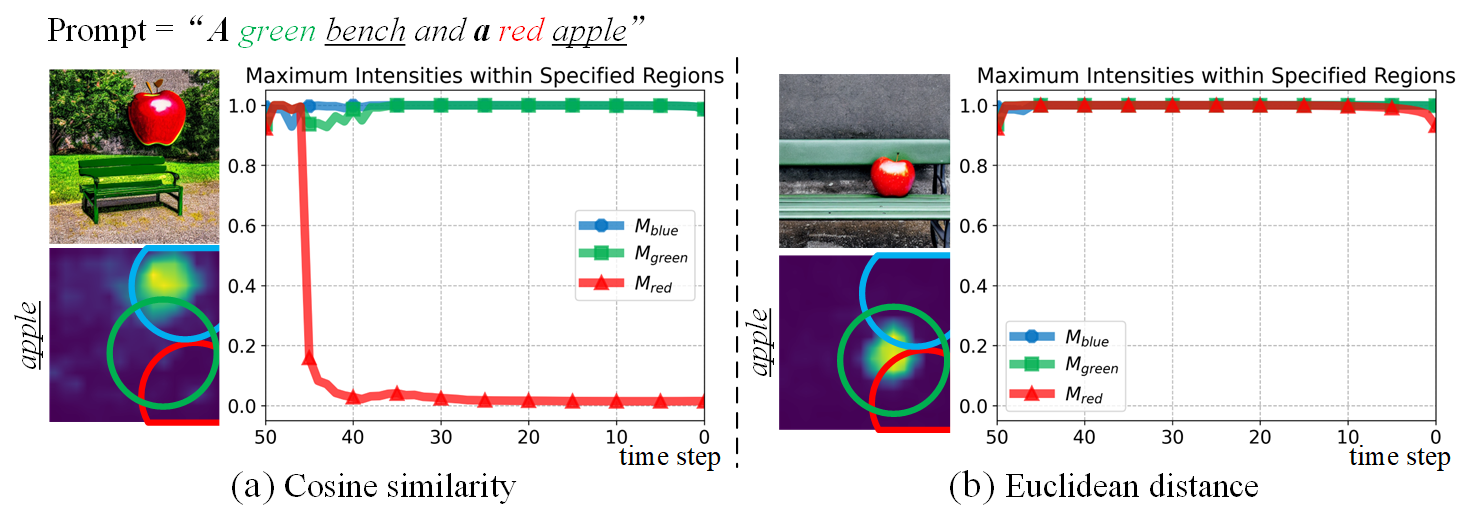}
	\caption{Comparison of (a) Cos+Cos and (b) Euc+Cos under case \textbf{C}, including generated images, corresponding cross-attention maps and the temporal variation of maximum activation values ($M_{b}$, $M_{g}$ and $M_{r}$) within each grouping region during denoising. Note that both (a) and (b) are based on identical initial noise and the grouping regions identified at the initial timestep.
	}
	\label{fig:agg_compare_statistic}
\end{figure}
\vspace{-0.3cm}

\section{Experiments}
\subsection{Experimental Details}
\noindent{\textbf{Model Choice.}} For fair comparison with prior works, all experimental results are obtained  based on SD v1.4 \cite{sd}. To evaluate the generalization of our AI-T2I across the diffusion models with varied architectures, we further integrate our AI-T2I into other diffusion models, \emph{e.g.}, SD v2.1, XL, and v3.5. 

\noindent\textbf{Datasets.} We experimentally evaluate AI-T2I over two typical text-to-image synthesis datasets, including: 1) \textit{AnE dataset} \cite{attend} comprises three types of text prompts: (i) “a [animalA] and a [animalB]”, (ii) “a [animal] and a [color][object]”, and (iii) “a [colorA][objectA] and a [colorB][objectB]”, where these text prompts consist of combinations between 12 animals and 12 object items with 11 colors; (2) \textit{T2I-CompBench++ dataset} \cite{t2ibench} comprises compositional text prompts evaluating complex compositions, and we use \textit{shape} text prompts. Notably, we report the experimental results by computing mean ± standard deviation with varied random seeds in Tab. \ref{tab:t2i_dataset}.

\noindent\textbf{Implementation Details.} Following ~\cite{attend,astar,conform}, the denoising process consists of 50 timesteps in total. We perform the overall optimization objective Eq. \ref{eq:overall_loss} to achieve precise text-to-image alignment during the initial 25 timesteps. For hyperparameters, ${\lambda _1}$, ${\lambda _2}$, ${\lambda _3}$ and ${\lambda_4}$ in Eq. \ref{eq:overall_loss} are set as 1.25, 2, 0.25 and 0.75. Regarding ${\cal L}_{agg-sub}$ (Eq. \ref{eq:agg_sub}), the number of grouping regions $N^c$ is set to $3$ with $r=5$ for objects, whereas for animals, it is $N^c=2$ with $r$ linearly increasing from 2 to 8 over the denoising process. For ${\cal L}_{agg-attr}$, $N^c=3$ with $r=6$.

\begin{figure}[t]
	\centering
	\includegraphics[width=0.8\linewidth]{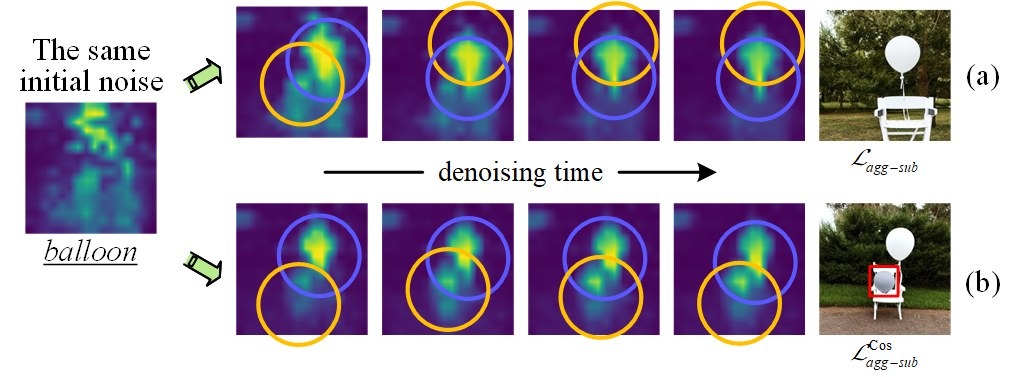}
	\caption{(a) The aggregation loss with the Euclidean distance. (b) A failure case of the aggregation loss with the cosine similarity.}
	\label{fig:cos_failure}
	\vspace{-3.5mm}
\end{figure}
\subsection{Distance Metric Comparison for Aggregation and Isolation Losses}
\label{sec:fur_inves}
Recalling Sec. \ref{sec:method}, the Euclidean distance is incorporated into the aggregation and isolation losses to address intra-token scattering and inter-token overlap. To shed more light on Euclidean distance, we perform validation experiments for the following cases: \textbf{A}: AI-T2I \emph{w/o} (${\cal L}_{iso}+{\cal L}_{agg-attr}$), \textbf{B}: AI-T2I \emph{w/o} (${\cal L}_{agg-sub}+{\cal L}_{agg-attr}$), and \textbf{C}: AI-T2I \emph{w/o} ${\cal L}_{agg-attr}$, in which we replace the Euclidean distance (Euc) in the aggregation and isolation losses with the cosine similarity (Cos), yielding several alternatives such as Cos+Cos and Euc+Cos. In such cases, ${\cal L}_{agg - sub}$ and ${\cal L}_{iso}$ are transformed into ${{\cal L}_{agg - sub}^{Cos}}$ and ${{\cal L}_{iso}^{Cos}}$, formulated as   
\begin{small}
\begin{equation}
	\begin{aligned}
			{{\cal L}_{agg - sub}^{Cos}} &= \sum\limits_{\scriptstyle i,k = 1, i \ne k\hfill}^{{N^c}} (1-{Cos(a _{t(i)}^m, a _{t(k)}^m)})
			\\&= \sum\limits_{\scriptstyle i,k = 1, i \ne k\hfill}^{{N^c}} (1-\frac{{\overrightarrow a _{t(i)}^m \cdot \overrightarrow a _{t(k)}^m}}{{\left\| {\overrightarrow a _{t(i)}^m} \right\| \cdot \left\| {\overrightarrow a _{t(k)}^m} \right\|}}), \\
			{{\cal L}_{iso}^{Cos}} &= Cos(A_t^m,A_t^n)   \\
			& = \frac{{\overrightarrow A _{t}^m \cdot \overrightarrow A _{t}^n} }{{\left\| {\overrightarrow A _{t}^m} \right\| \cdot \left\| {\overrightarrow A _{t}^n} \right\|}}, 
	\end{aligned}
\end{equation}
\end{small}
where $Cos(\cdot)$ denotes the cosine similarity function. $\overrightarrow a _{t(i)}^m$ indicates the vector representation of the $i$-th grouping region at the $t$-th timestep; $\overrightarrow A _{t}^n$ denotes the vectorized representation of $A_t^n$.
The results in Tab. \ref{tb:blip_cos_euc_compare_full} suggests that  Cos+Cos and Euc+Cos suffer from a large performance degradation (at most 0.015\%), which is attributed to the advantages of Euclidean distance in capturing spatial locations. In particular, Euc+Euc exhibits the obvious performance gains (at most 0.018\%) over Euc+Cos, confirming the \emph{importance} of Euclidean distance in the isolation loss, owing to the \emph{adversarial relationship} between the aggregation and isolation losses.

\begin{figure}[t]
	\centering
	\setlength{\abovecaptionskip}{0.2cm}
	\setlength{\belowcaptionskip}{-0.4cm}
	\includegraphics[width=1.0\columnwidth]{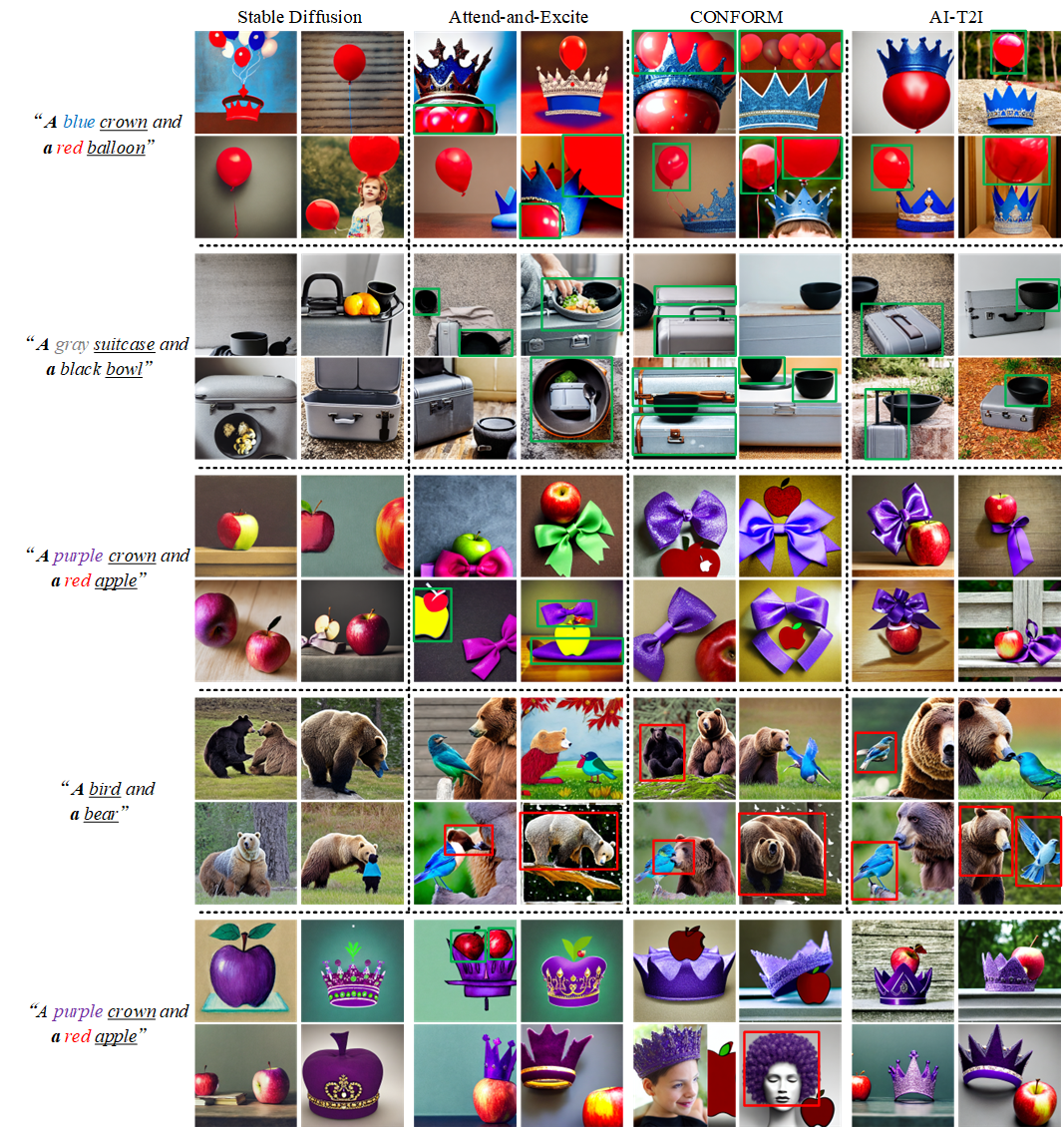}
	\caption{Qualitative comparison of AI-T2I with the state-of-the-art methods. For each prompt, four generated images are shown with the same set of seeds across all approaches. Our AI-T2I delivers the desirable subject generation (marked as the red and green boxes) against others.}
	\label{fig:qualitative}
\end{figure}

Additionally, Fig. \ref{fig:agg_compare_statistic} presents a visual comparison of the Cos+Cos and Euc+Cos configurations under Case C, both initialized from the same noise. The results include the generated images, their corresponding cross-attention maps, and the temporal evolution of the maximum activation values ($M_b$, $M_g$, $M_r$) within each grouping region during denoising. Given that both configurations recompute grouping regions at each timestep (causing their positions to diverge), we visualize and compare the regions identified at the initial step. This allows us to track how activations evolve under different metrics at fixed spatial locations. This controlled setup ensures that any differences in the final outputs are attributable solely to the choice of distance metric, not to stochastic variations in the denoising process. The statistical results demonstrate that cosine similarity settles intra-token scattering by minimizing the maximum intensity within the non-overlapped grouping
region, rather than inducing convergence of all regions like
Euclidean distance. It is observed that Cos+Cos receives a mislocated and unrealistic synthesized images despite the correct text-to-image alignment, owing to the \emph{failure} to pull together all the centers of intra-token activation regions. In contrast, Euc+Euc induces convergence of the centers across all grouping regions, yielding the desirable subject generation, which is consistent with \emph{our analysis} in Sec. \ref{sec:method}.

\begin{table}[t]
	\caption{Comparison of quantitative results with other state-of-the-art works on the T2I-CompBench++ dataset. The best and second-best results are reported with \textbf{boldface} and \underline{underline}, respectively.}
		\centering
		\scriptsize
		\begin{tabular}{c c c c}
			\toprule
			\textbf{Method} & \textbf{Full Sim. (\(\uparrow\))}  & \textbf{Min Sim. (\(\uparrow\))}&  \textbf{T-C Sim. (\(\uparrow\))} \\
			\midrule
			StableDiffusion\cite{sd} & 0.316 &0.246  & 0.719\\
			AnE\cite{attend} & 0.322 & 0.256 & 0.732  \\
			PredicatedDiff\cite{predicated} & 0.319 & 0.252 & 0.724 \\
			Self-cross\cite{2025selfcross} & 0.324 & 0.260 & 0.724 \\
			\midrule  
			\rowcolor{c1!40}
			\textbf{AI-T2I (Ours)} & \makecell{\textbf{0.327} \\ \textcolor[RGB]{142,68,198}{$ \pm 5.30 \times {10^{ - 4}}$}
			}
			& 
			\makecell{\textbf{0.262} \\ \textcolor[RGB]{142,68,198}{$ \pm 5.40 \times {10^{ - 4}}$}} & \makecell{\textbf{0.738} \\ \textcolor[RGB]{142,68,198}{$ \pm 3.15 \times {10^{ - 3}}$}} \\
			\bottomrule
		\end{tabular}
	
	\label{tab:t2i_dataset}
	\vspace{-3.5mm}
\end{table}

\begin{figure}[t]
	\centering
	\includegraphics[width=\linewidth]{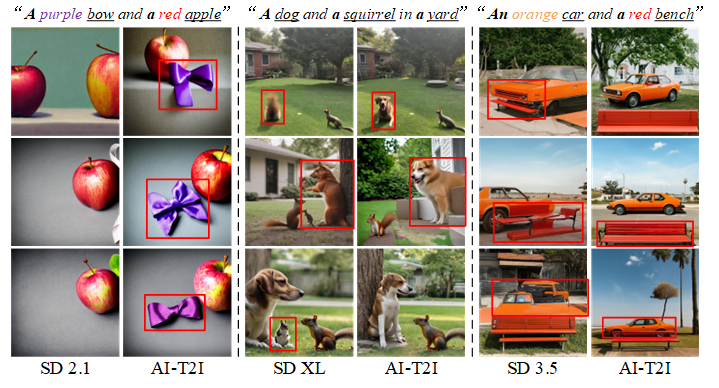}
	\caption{The visual results of AI-T2I on multiple SD versions, confirming the effectiveness of AI-T2I when applied to varied Stable Diffusion versions.}
	\label{fig:extensive_version}
\end{figure}

In particular, we also visualize some failure cases of the aggregation loss with the cosine similarity. Fig. \ref{fig:cos_failure} illustrates that the Euclidean distance admits an obvious separation between two high-intensity regions in Fig. \ref{fig:cos_failure}(a), while the cosine similarity fails in Fig. \ref{fig:cos_failure}(b); the reason is that the cosine similarity focuses solely on aligning corresponding elements within the grouping regions, rather than aggregating them. In contrast, the aggregation loss with the Euclidean distance achieves finer-grained aggregation of scattering activations by leveraging centroids as intermediaries. Such failure case verifies the advantages of the Euclidean distance over the cosine similarity for the aggregation loss.

\subsection{Comparison with the State-of-the-art works}

To verify the superiority of AI-T2I, we compare it with the typical text-to-image approaches, including: Stable Diffusion \cite{sd} serves as the baseline; Structure \cite{structure}, EBCA \cite{park2024energy}, AnE \cite{attend}, SG \cite{linguistic}, PredicatedDiff \cite{predicated}, Atten\&Regulation \cite{zhang2025enhancing} and Self-cross \cite{2025selfcross} focus on inter-token relationship; CONFORM\cite{conform} relies on the activations from the previous denoising timesteps to address the intra-token scattering.
Following \cite{attend}, our AI-T2I is compared via several metrics: \emph{Full Prompt Similarity} (\emph{Full. Sim.}) computes the average CLIP \cite{clip} cosine similarity score between the text prompt and generated images; \emph{Minimum Object Similarity} (\emph{Min. Sim.}) measures the CLIP similarity between each subject sub-prompt and generated images, to measure whether each subject in the prompt has been successfully generated; and \emph{Text-Caption Similarity} (\emph{T-C Sim.}) adopts a pre-trained BLIP image-captioning model \cite{li2022blip} to generate captions for each image and computes the average CLIP similarity between the original prompt and BLIP descriptions. In particular, we generate 64 images for each prompt.

\begin{table}[t]
	\caption{Benchmarking of computational overhead (TFLOPs, GPU memory, and inference time) for $512\times512$ image generation. Results show AI-T2I achieves a favorable balance between added cost and alignment performance. $\downarrow$: Lower is better.}
		\centering
		\scriptsize
		\begin{tabular}{c c c c}
			\toprule
			\textbf{Method} & \textbf{TFLOPs ($\downarrow$)}  & \textbf{GPU Memory (GB) ($\downarrow$)} & \textbf{Time (s) ($\downarrow$)} \\
			\midrule
			StableDiffusion\cite{sd} & 34.55 &7.98  & 8.92\\
			AnE\cite{attend} & 51.83 & 13.98 & 16.74  \\
			PredicatedDiff\cite{predicated} & 51.83 & 15.46 & 16.81 \\
			Self-cross\cite{2025selfcross} & 51.83 &13.80 & 24.53 \\
			\midrule  
			\rowcolor{c1!40}\textbf{AI-T2I (Ours)} & 51.83 & 14.38 & 16.41
			\\
			
			\bottomrule
		\end{tabular}
	
	\label{tab:computational_cost}
\end{table}	

\begin{table}[t]
	\centering
	\scriptsize
	\caption{Performance of AI‑T2I on diffusion architectures beyond Stable Diffusion. Results on FLUX.1-dev, HunyuanDiT and Qwen‑Image, evaluated with  \emph{Full Sim.}, \emph{Min. Sim.} and \emph{T-C Sim.} metrics. $\uparrow$: Higher is better.}
	\begin{tabular}{l c c c} 
		\toprule
		
		\textbf{Models} & \textbf{Full Sim. (\(\uparrow\))} & \textbf{Min. Sim. (\(\uparrow\))} & \textbf{T-C Sim. (\(\uparrow\))}\\
		\midrule
		
		\rowcolor{gray!30}
		\multicolumn{4}{@{}l}{\textbf{A}: FLUX.1-dev} \\
		Baseline \cite{flux}    &0.361   &0.270  &0.861  \\
		AI-T2I    &0.367 &0.275  &0.870 \\             
		\midrule
		
		\rowcolor{gray!30}
		\multicolumn{4}{@{}l}{\textbf{B}: HunyuanDiT} \\
		Baseline \cite{hunyuandit2024}    &0.370  &0.280  &0.852  \\
		AI-T2I    &0.374  &0.284  &0.863  \\  
		\midrule
		\rowcolor{gray!30}
		\multicolumn{4}{@{}l}{\textbf{C}: Qwen-Image} \\
		Baseline \cite{qwenimage2025}   &0.380  &0.281  &0.895  \\
		AI-T2I    &0.379  &0.281  &0.897  \\                 		
		
		\bottomrule
	\end{tabular}
	\label{tb:other_models}
	
\end{table}

\noindent \textbf{Qualitative and Quantitative Results.} The qualitative results in Fig. \ref{fig:qualitative}, apart from quantitative results in Tab. \ref{tb:metric_results} and Tab. \ref{tab:t2i_dataset}  summarize our findings below: 
1) AI-T2I enjoys a larger similarity scores (at most 0.116), \emph{i.e.}, \emph{Full. Sim.}, \emph{Min. Sim.} and \emph{T-C Sim.}, than other competitors, apart from the excellent text-to-image alignment (the col 7 and 8 of Fig. \ref{fig:qualitative}), \emph{confirming} that AI-T2I effectively addresses intra-token scattering and inter-token overlap. 
2) Compared to AI-T2I, CONFORM fails to overcome the intra-token scattering (\emph{e.g.}, multiple red balloons in the col 5 and 6 of Fig. \ref{fig:qualitative}), owing to great reliance on the activations from the previous denoising steps, which, in turn, verifies the \emph{necessity} of our aggregation loss (Eq. \ref{eq:agg_sub}) focusing on the activations from the current denoising step in Sec. \ref{sec:agg_loss}. 
3) AnE, SG, PredicatedDiff and Atten\&Regulation focus on inter-token relationship, they bear an obvious performance loss and the inevitable subject overlap (the col 3 and 4 of Fig. \ref{fig:qualitative}), verifying the \emph{advantages} of exploiting the isolation loss with Euclidean distance to address inter-token overlap over others. 
Fig. \ref{fig:extensive_version} presents the results of AI-T2I on multiple SD versions, \emph{e.g.}, SD v2.1, SD XL 1.0, and SD 3.5, showing that AI-T2I remains effective when applied to different Stable Diffusion versions. \\

\subsection{Computational Overhead Analysis}
	To evaluate the computational cost of our gradient-based latent optimization, we profile the inference of $512\times512$ image generation over 51 denoising steps. As shown in Tab. \ref{tab:computational_cost}, compared to the baseline SD without optimization, AI‑T2I increases FLOPs by about 50\%, wall-clock denoising time by 84\%, and peak GPU memory by 80\%. These overheads arise primarily from the extra UNet forward pass required for gradient computation, while the loss computations themselves are negligible. Therefore, despite different loss functions, all compared state‑of‑the‑art optimization-based methods have identical FLOPs (51.83 T). Notably, AI‑T2I receives slightly lower denoising times  than compared methods (16.41 s vs. 16.74 s, 16.81 s and 24.53 s, respectively). This is because the compared optimization-based methods employ an inner‑loop iterative refinement mechanism at certain timesteps, which performs multiple gradient‑based updates (up to 20 iterations) within a single denoising step, thus ensuring that attention thresholds are met. In contrast, AI‑T2I performs a single gradient update per denoising step, yielding faster inference speed while maintaining high performance, benefiting from the well‑designed aggregation‑and‑isolation losses. The above results confirm that our proposed losses introduce moderate and manageable overhead while delivering desirable text‑to‑image alignment.

\begin{figure}[t]
	\centering
	\setlength{\abovecaptionskip}{0.2cm}
	\setlength{\belowcaptionskip}{-0.4cm}
	\includegraphics[width=1.0\columnwidth]{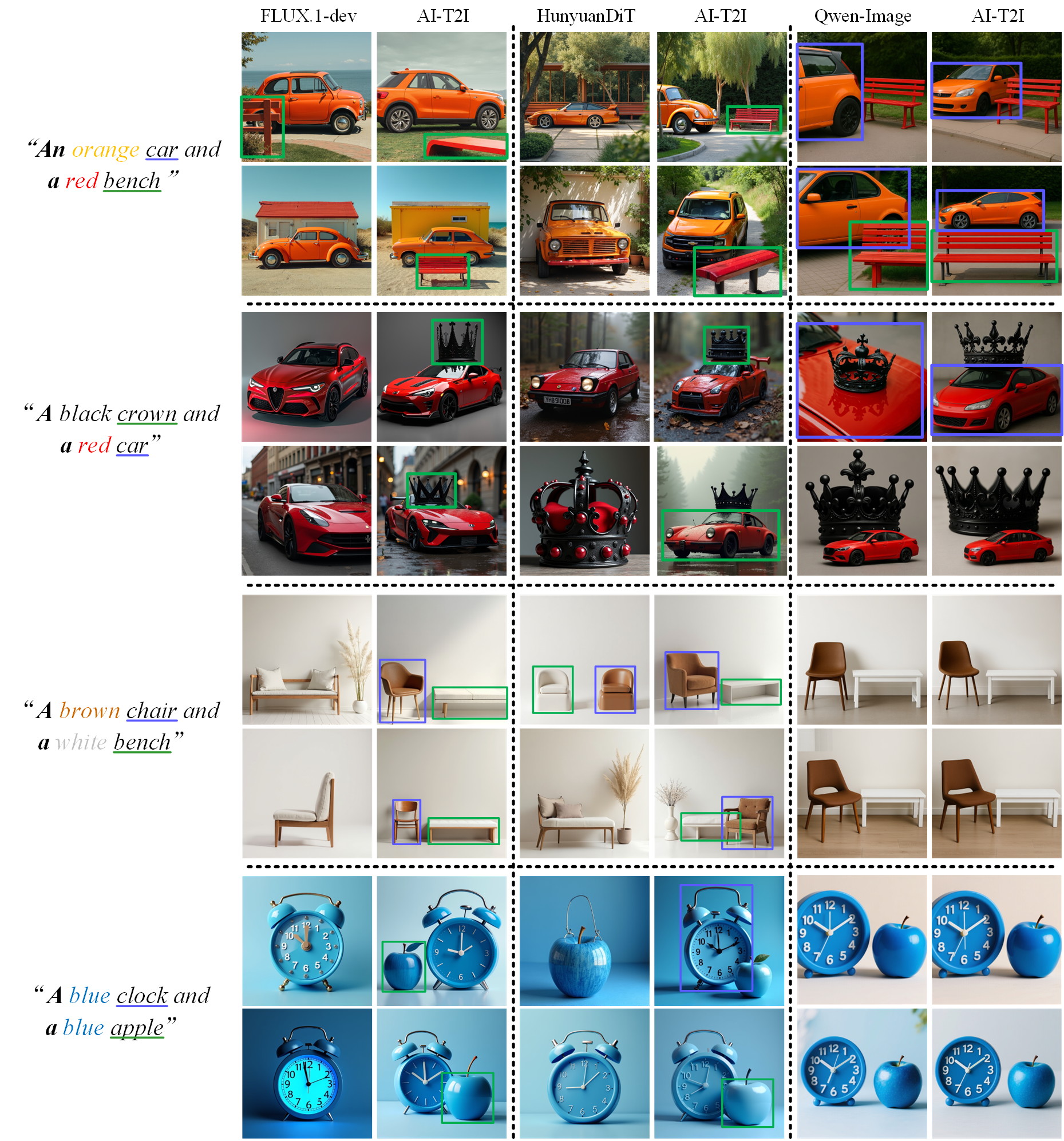}
	\caption{Qualitative comparison of transformer-based models (i.e., FLUX.1-dev, HunyuanDiT, Qwen-Image) and their AI-T2I-enhanced versions. For each prompt, we show representative samples generated with identical random seeds. AI-T2I consistently improves multi-subjects generation and attribute binding (marked as the purple and green boxes).}
	\label{fig:dit_qualitative}
\end{figure}	

\begin{figure}[t]
	\centering
	\includegraphics[width=0.95\linewidth]{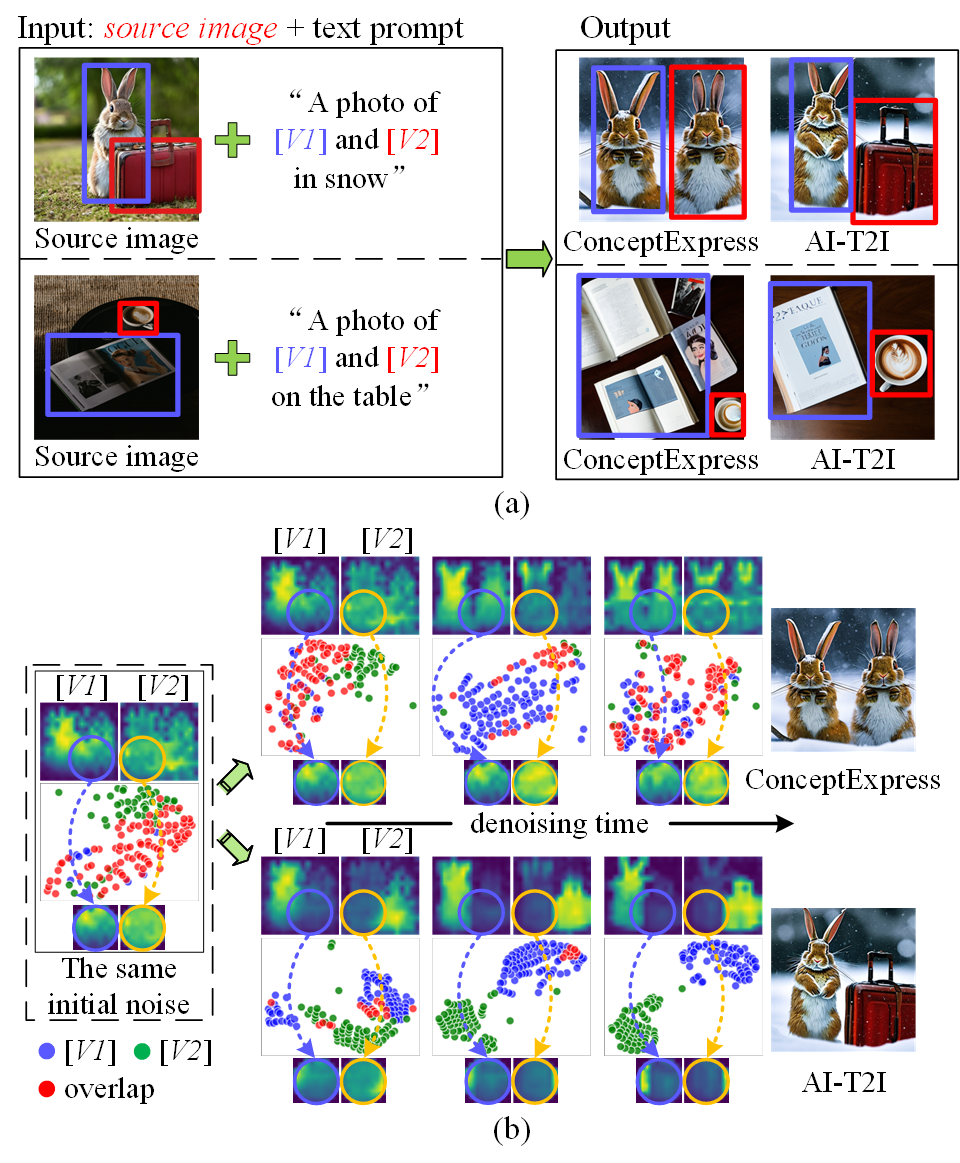}
	\caption{Extension of  AI-T2I on the personalized generation task with additional inputs beyond text prompts. (a) Personalized generation \cite{hao2024conceptexpress}, where [\textit{Vi}] is a visual concept from the source image. (b) The comparison of denoising process between the baseline and AI-T2I. The colored dots represent activations of subject attention maps following t-SNE projection.}
	\label{fig:extensive_task_explain}
\end{figure}

\subsection{Generalization to Other Diffusion Architectures}

To further examine the generalizability of the proposed AI-T2I beyond the SD family, we conduct experiments on diffusion backbones with different architectures, involving FLUX.1-dev \cite{flux}, HunyuanDiT \cite{hunyuandit2024} and Qwen-Image \cite{qwenimage2025}. We evaluate AI-T2I on the Objects-Objects subset of the AnE dataset (images generated at $512\times512$ resolution due to computational constraints). \\
\newpage
	\noindent \textbf{Experimental Details.} The FLUX.1-dev model comprises 19 joint transformer blocks and 38 single transformer blocks. We extract cross-attention maps from 6 selected joint transformer blocks. For Qwen‑Image, we sample cross‑attention maps from 30 out of its 60 transformer blocks. Notably, FLUX.1-dev and Qwen-Image employ a unified joint‑attention computation where image and text tokens participate jointly, instead of computing self-attention and cross-attention separately as in SD 1.4. The HunyuanDiT model comprises 40 transformer blocks, from which we extract maps at 30 selected blocks. As this model concatenates the token embeddings from two text encoders (CLIP and T5) to project the Key and Value matrices, the resulting cross-attention maps naturally contain two distinct parts, each aligned with one text encoder. Consequently, we compute the loss functions separately over these two corresponding segments. The weights of these two loss terms are determined by the following formulas:
	\begin{equation}
		\begin{aligned}
			\tau &= \max({A}_{\text{clip}}) + \max({A}_{\text{t5}}) \\
			\lambda_{\text{clip}} &= \frac{\max({A}_{\text{clip}})}\tau \\
			\lambda_{\text{t5}} &= \frac{\max({A}_{\text{t5}})}\tau
		\end{aligned}
		\label{eq:weight_calculation}
	\end{equation}
	where $A_{\text{clip}}$ and $A_{\text{t5}}$ denote the cross‑attention maps obtained from the CLIP and T5 text encoders, respectively. Then the final loss of HunyuanDiT is ${\cal L} = \lambda_{\text{clip}}{{\cal L}_{clip}} + \lambda_{\text{t5}}{{\cal L}_{t5}}$. Furthermore, since the size of attention maps in DiT‑based models are typically larger than the commonly used $16\times16$ scale in UNet architectures, we proportionally increase the grouping radius $r$ in the aggregation loss to maintain compatibility with the attention-map scale.\\

	\begin{figure}[t]
		\centering
		\includegraphics[width=0.95\linewidth]{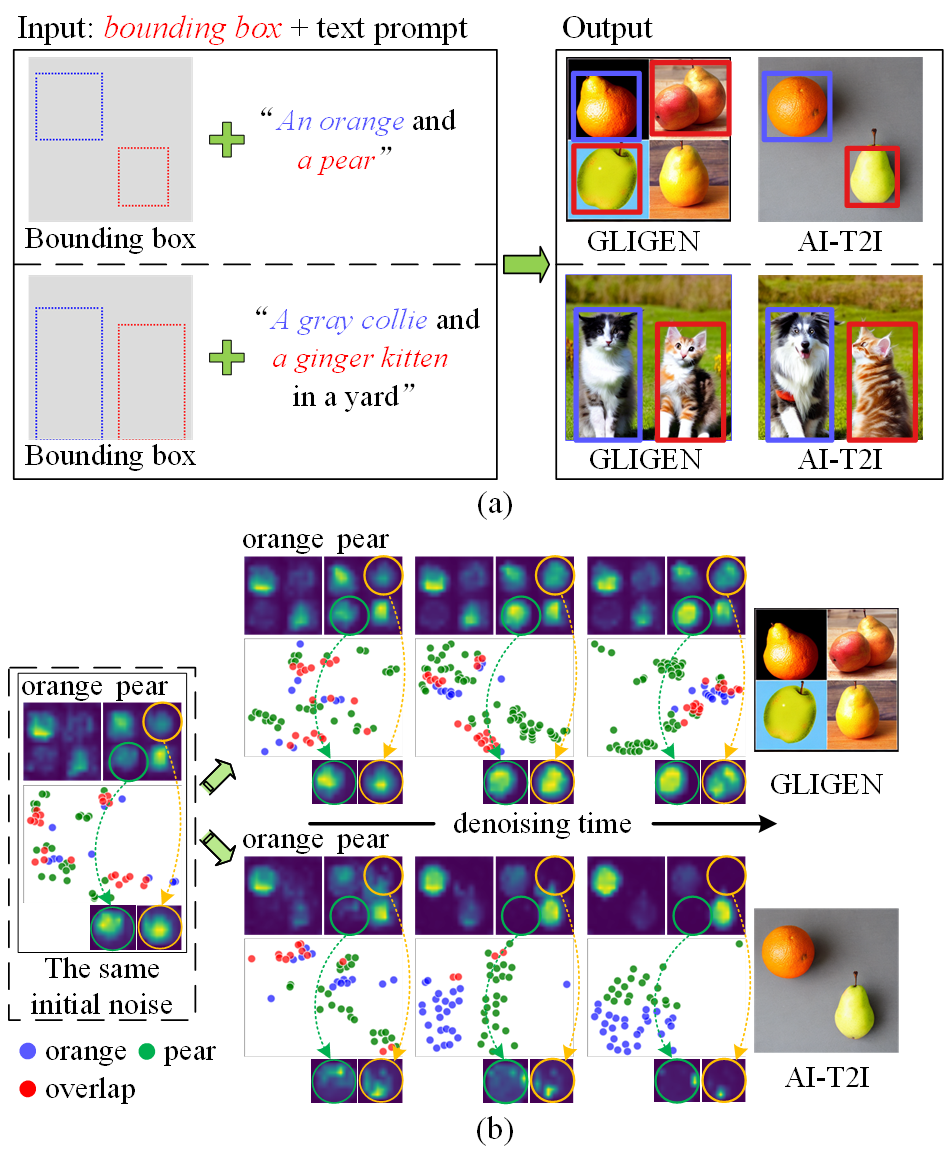}
		\caption{Extension of  AI-T2I on the controllable layout generation task with additional inputs beyond text prompts. (a) Controllable layout generation with bounding boxes \cite{li2023gligen}. (b) The comparison of denoising process between the baseline and AI-T2I. The colored dots represent activations of subject attention maps following t-SNE projection.}
		\label{fig:extensive_task}
	\end{figure}

	\noindent \textbf{Qualitative and Quantitative Results.}
	As summarized in Tab. \ref{tb:other_models}, incorporating AI-T2I brings consistent improvements across all metrics on both FLUX.1-dev and HunyuanDiT. On Qwen-Image, whose baseline alignment is already strong, AI-T2I maintains comparable performance without degradation. These results confirm that AI-T2I is not tailored to a specific diffusion architecture; it effectively enhances (or preserves) text-to-image alignment across varied generative backbones, demonstrating plug-and-play versatility. Additionally, Fig. \ref{fig:dit_qualitative} further exhibits the quantitative results. It is observed that, when AI-T2I is applied to these transformer-based diffusion models, visible enhancements in prompt fidelity are obtained. For instance, in challenging multi-subject prompts, FLUX.1-dev and HunyuanDiT sometimes fail to generate both subjects or mix their attributes (e.g., the \enquote{chair} in row 5, column 1; the \enquote{apple} in row 7, column 1), whereas AI-T2I successfully produces both subjects with clear boundaries and correct attribute binding (e.g., the \enquote{chair} in row 5, column 2; the \enquote{apple} in row 7, column 2). Even for Qwen-Image, which already demonstrates strong native alignment, AI-T2I further refines fine-grained details. Although the baseline often generates both subjects, one of them is not fully completed (e.g., the \enquote{car} in rows 1$\sim$3, column 5). In contrast, with AI-T2I, all mentioned subjects are distinctly and completely present (e.g., the \enquote{car} in row 1$\sim$3, column 6). Besides, AI-T2I can still maintain the baseline's performance even though the baseline already generates complete subjects (e.g., rows 4$\sim$8, column 6). These results confirm that AI-T2I serves as a lightweight, training-free enhancer that improves text-to-image alignment in challenging multi-subject scenarios across diverse architectures.\\

\begin{table}[t]
	\small
	\centering
	,	\caption{Quantitative Results for the extensive tasks, \emph{i.e.}, Controllable Layout Generation (IoU on COCO-Stuff) and Personalized Generation (SIM on ConceptExpress).} 
	\scalebox{0.9}{
			
			\begin{tabular}{c c c} 
				\toprule  
				\textbf{Controllable Layout Generation} &\textbf{Layout Data} & \textbf{IoU} (\(\uparrow\))  \\
				\midrule 
				GLIGEN\cite{li2023gligen} & COCO-Stuff &  0.58  \\
				\rowcolor{c1!40} GLIGEN + \textbf{AI-T2I (Ours)} & COCO-Stuff & \textbf{0.63}  \\
				
				\midrule
				\midrule
				\textbf{Personalized Generation} & \textbf{Personalized Data} & \textbf{SIM} (\(\uparrow\))\\
				\midrule
				ConceptExpress \cite{hao2024conceptexpress} & ConceptExpress &0.745 \\
				\rowcolor{c1!40} ConceptExpress + \textbf{AI-T2I (Ours)} & ConceptExpress &\textbf{0.763} \\					
				
				\bottomrule
				
			\end{tabular}
		}
		
		\label{tab:extensive_tasks}
	\end{table}
	
\begin{figure}[t]
	\centering
	\includegraphics[width=\linewidth]{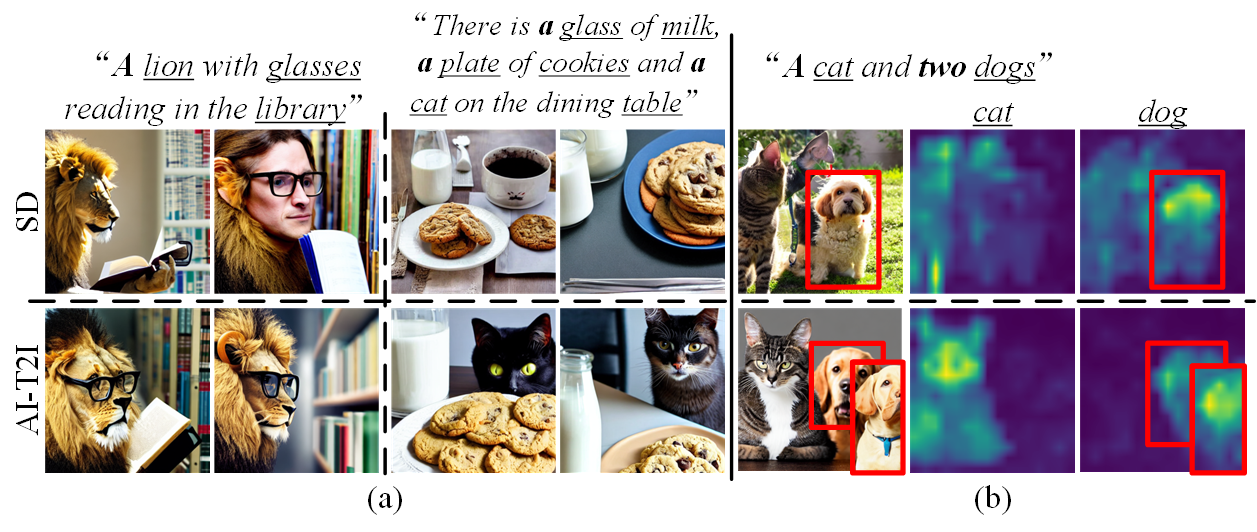}
	\caption{Visual analysis on multi-subject scenarios of AI-T2I, involving multiple (a) different and (b) identical subjects.}
	\label{fig:multi_concepts}
\end{figure}

\noindent \textbf{Positioning AI-T2I Relative to Large-Scale Generative Models.}
	The preceding experiments utilize recent, large-scale models (e.g., FLUX, HunyuanDiT, and Qwen-Image) as baselines. While such models achieve high fidelity through massive scale and training, AI‑T2I operates under a complementary paradigm as a training‑free, model‑agnostic enhancer rather than a competing generative system. This distinction is evident in several key aspects: AI‑T2I requires no retraining, thereby avoiding the substantial cost of scaling parameters or data; it generalizes across distinct architectures to consistently improve alignment; and it is task‑general, enabling seamless integration into various downstream applications such as layout‑controlled generation. Thus, AI‑T2I provides a portable solution that extends the alignment capability of pre‑existing models where retraining is impractical, without requiring architectural changes. It is worth noting that Qwen‑Image, as the largest and most recent model among the three evaluated in our experiments, already exhibits strong native text‑image alignment capability. Although its quantitative metrics do not substantially surpass those of the other two baselines, qualitative observation reveals that the images it generates consistently conform to the textual conditions. This indicates that through large-scale training, Qwen-Image has significantly improved native alignment. However, the consistent gains observed in FLUX.1-dev and HunyuanDiT (Tab. \ref{tb:other_models}) demonstrate that AI-T2I remains a potent enhancer for most modern architectures. Even for the most advanced models, AI-T2I serves as a safety guard to further ensure alignment without any additional inference-side architectural changes. This precisely indicates that AI‑T2I is a training‑free, plug‑and‑play enhancer that complements rather than competes with such models. It provides a lightweight solution when retraining is infeasible.
	
	\begin{figure}[t]
		\centering
		\includegraphics[width=0.8\linewidth]{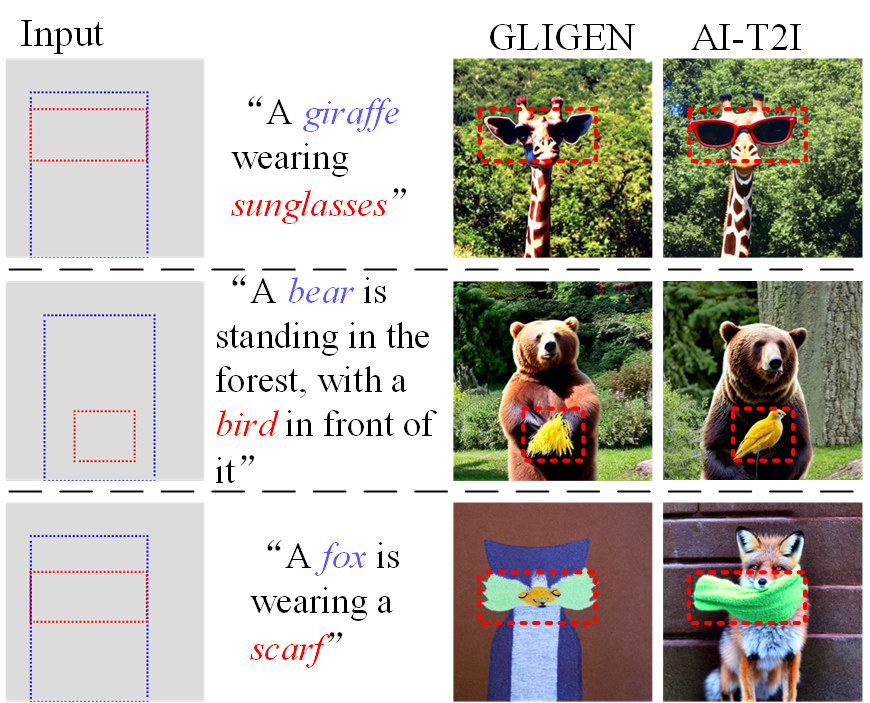}
		\caption{Qualitative comparison between GLIGEN and AI-T2I on complex interaction prompts with overlapping bounding boxes. GLIGEN often fails to separate overlapping subjects, leading to missing or merged objects. In contrast, AI-T2I successfully generates both subjects with clear structural boundaries by aggregating intra-token and isolating inter-token activations (marked as the red boxes).}
		\label{fig:overlap_box}
	\end{figure}

\subsection{Extension to Other Tasks}
To validate the generalization of AI-T2I on other tasks conditioned on the inputs beyond
text prompts, we further extend our AI-T2I to the controllable layout generation and personalized generation tasks. 

\noindent \textbf{Controllable Layout Generation}. For the controllable layout generation task,  the generative model produces the images in which the semantic content aligns with text prompts and the layout conforms to user-specified \textit{bounding boxes} \cite{li2023gligen,zheng2025semantic,taghipour2025box}. As illustrated in Fig.~\ref{fig:extensive_task}, AI-T2I is exploited to consolidate the bounding boxes constraints by using the aggregation and isolation losses to steer the spatial coordinates of subject-token activations to fall inside the prescribed boxes. Fig.~\ref{fig:extensive_task}(b) compares the denoising processes of the baselines and AI-T2I with the same initial noise. For example, the GLIGEN produces redundant \textit{pears} outside the bounding boxes, as shown in the first row of Fig.~\ref{fig:extensive_task}(b), where the cross-attention maps exhibit significant inter-token overlap and intra-token scattering issues. In contrast, AI-T2I yields the images with the semantics that adhere to the prompt and the layouts that respect the bounding-box constraints. Besides, the t-SNE maps of subject-token activations exhibit a effective mitigation of inter-token overlap issue with our AI-T2I. Furthermore, by normalizing token activations within grouping regions and visualizing the resulting patterns (green and orange circles), we confirm the effectiveness of AI-T2I to aggregate scattered activations.

To further evaluate the effectiveness of AI-T2I under complex interaction scenarios, we conduct additional experiments on prompts involving spatially overlapping subjects. In these cases, the input bounding boxes naturally overlap, making layout-guided methods prone to failure. As shown in Fig. \ref{fig:overlap_box}, GLIGEN often produces merged or missing subjects due to ambiguous spatial guidance (e.g., the \enquote{sunglasses} in row 1). In contrast, AI-T2I consistently generates both subjects with clear structural boundaries. These results validate that AI-T2I is not limited to spatially separated objects but also excels in handling semantic entanglement when subjects overlap spatially.

\begin{table}[t]
	\small
	\centering
	\caption{Ablation study about varied components constituting AI-T2I by measuring text-caption similarity (T-C Sim.). The best results are reported with \textbf{boldface}. *: Since $L_{agg-attr}$ is not applicable to the Animal–Animal subset, the actual configuration of case D is identical to that of case C in this subset, resulting in the same performance.}
	\begin{tabular}{c c c c} 
		\toprule
		\textbf{Cases} & \textbf{Object-Object} &  \textbf{Animal-Object} & \textbf{Animal-Animal}  \\
		\midrule
		\rowcolor{gray!30}Baseline & 0.765  & 0.793   & 0.767   \\
		\textbf{A}& 0.822 & 0.843   & 0.804 \\
		\textbf{B} & 0.823  & 0.844    & 0.833    \\
		\textbf{C} & 0.828   & 0.847   & \textbf{0.834}  \\
		\rowcolor{c1!40} 
		\textbf{D} & \textbf{0.838}   & \textbf{0.849}   & \textbf{0.834}\rlap{\textbf{*}}  \\
		\bottomrule
	\end{tabular}
	
	\label{tb:ablation_study_blip}
\end{table}

\begin{figure}[t]
	\centering
	\includegraphics[width=0.9\linewidth]{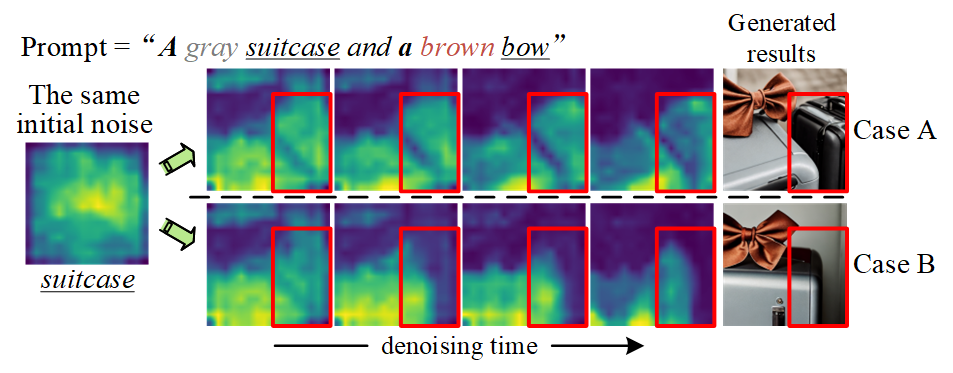}
	\caption{Additional ablation study about ${\cal L}_{agg-sub}$ for aggregation loss, with visual results of generated images and cross-attention maps for \textit{suitcase} across denoising timesteps. AI-T2I is capable of achieving noticeable subject aggregation.}
	\label{fig:aggregate}
\end{figure}

\noindent \textbf{Personalized Generation Task}. For the personalized generation task, \textit{special text tokens} are used to capture visual concepts from source images and  inject those concepts into diverse target scenes \cite{hao2024conceptexpress,xu2024sgdm}. AI-T2I is  exploited to improve the text-to-image alignment by reducing inter-token overlap and mitigating intra-token scattering of the \textit{special token} activations, as illustrated in Fig.~\ref{fig:extensive_task_explain}.

Tab. \ref{tab:extensive_tasks} summarizes the quantitative results for both controllable layout generation and personalized generation. For controllable layout generation, we followed the experimental setup of \cite{xie2023boxdiff} on COCO-Stuff \cite{2018COCO}, where object bounding boxes were detected using YOLOv4 \cite{2020YOLOv4} and the synthesis accuracy was assessed via Intersection over Union (IoU); while for personalized generation, the experiments were conducted on the ConceptExpress dataset \cite{hao2024conceptexpress}, meantime the performance is evaluated using CLIP-based similarity (denoted as SIM) between each concept in the training image and the corresponding concept-specific generated images. The results in Tab. \ref{tab:extensive_tasks} show that the incorporation of AI-T2I leads to consistent performance gains in both tasks, underscoring its effectiveness as a plug-and-play module to enhance existing models across various tasks.

	\begin{figure}[t]
	\centering
	\includegraphics[width=0.95\linewidth]{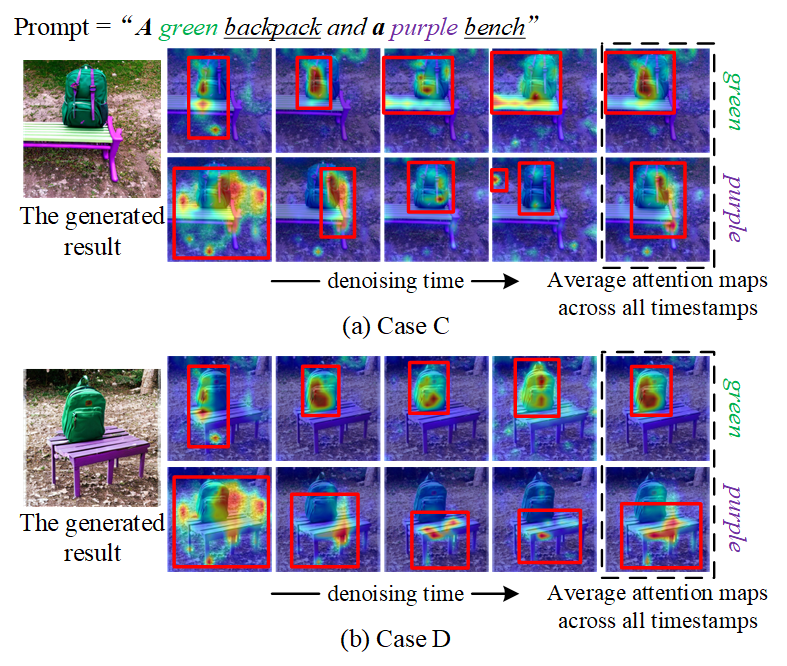}
	\caption{Ablation study about ${\cal L}_{agg-attr}$ for aggregation loss, with visual results of generated images and corresponding attention maps for attribute tokens across denoising timesteps.
	}
	\label{fig:adj}
\end{figure}

\subsection{Extension to Multiple Subjects}

To further show the generalization of AI-T2I, we extend AI-T2I to multi-subject scenarios by exploiting two types of multi-subject prompts, involving multiple either different or identical subjects. Specifically, we perform the visual analysis with the prompts as ``\emph{\textbf{A} \underline{lion} with \underline{glasses} reading in the \underline{library}}" and ``\emph{There is \textbf{a} \underline{glass} of \underline{milk}, \textbf{a} \underline{plate} of \underline{cookies} and \textbf{a} \underline{cat} on the dinner \underline{table}}" (see Fig. \ref{fig:multi_concepts}(a)), apart from ``\emph{\textbf{A} \underline{cat} and \textbf{two} \underline{dogs}}" (see Fig. \ref{fig:multi_concepts}(b)).  Fig. \ref{fig:multi_concepts}(a) illustrates that previous approaches (SD) either miss important concepts (\emph{e.g.}, glasses in the col 1 of Fig. \ref{fig:multi_concepts}(a)) or encounter improper mixture of concepts (\emph{e.g.}, lion and glasses in the col 2 of Fig. \ref{fig:multi_concepts}(a)) compared to  AI-T2I; while for Fig. \ref{fig:multi_concepts}(b), SD fails to distinguish between multiple identical subjects (\emph{e.g.}, two dogs in the col 1 of Fig. \ref{fig:multi_concepts}(b)). The above fact further verifies \emph{our proposal of simultaneously addressing intra-token scattering and inter-token overlap by aggregating-and-isolating intra-token and inter-token activations}.  
	
\subsection{Ablation Study}

\noindent \textbf{Is each component of AI-T2I essential?}
	To verify the effectiveness of varied components constituting AI-T2I, we perform the ablation experiments from the following cases: \textbf{A}: removing ${\cal L}_{iso}$ for all subsets (since the Animal–Animal subset contains no attribute tokens, $\mathcal{L}_{agg-attr}$ is omitted for that subset but retained for the others); \textbf{B}: removing both ${\cal L}_{agg-sub}$ and ${\cal L}_{agg-attr}$; \textbf{C}: removing ${\cal L}_{agg-attr}$; and \textbf{D}: the proposed AI-T2I. Tab. \ref{tb:ablation_study_blip} suggests that case \textbf{D} delivers great gains to other cases, which is \emph{evidence} that the collaboration between the aggregation and isolation loss succeeds in achieving precise text-to-image alignment. Besides, case \textbf{D} upgrades beyond case \textbf{B} and \textbf{C}, verifying the effectiveness of the aggregation loss in mitigating intra-token scattering. Notably, case \textbf{A} bears an obvious performance loss, since Euclidean distance is capable of pushing the isolation loss to alleviate inter-token overlap.
	Additionally, Fig. \ref{fig:adj} visualizes case \textbf{C} and \textbf{D} via generated images and corresponding attention maps for attribute tokens across denoising steps. We observe that case \textbf{C} suffers from the incorrect attribute binding (\emph{i.e.}, the green and purple) unlike case \textbf{D}, implying that, with ${\cal L}_{agg-attr}$, AI-T2I is \emph{capable} of achieving the attribute aggregation (Sec. \ref{sec:agg_loss}). 

\begin{figure}[t]
	\centering
	\includegraphics[width=\linewidth]{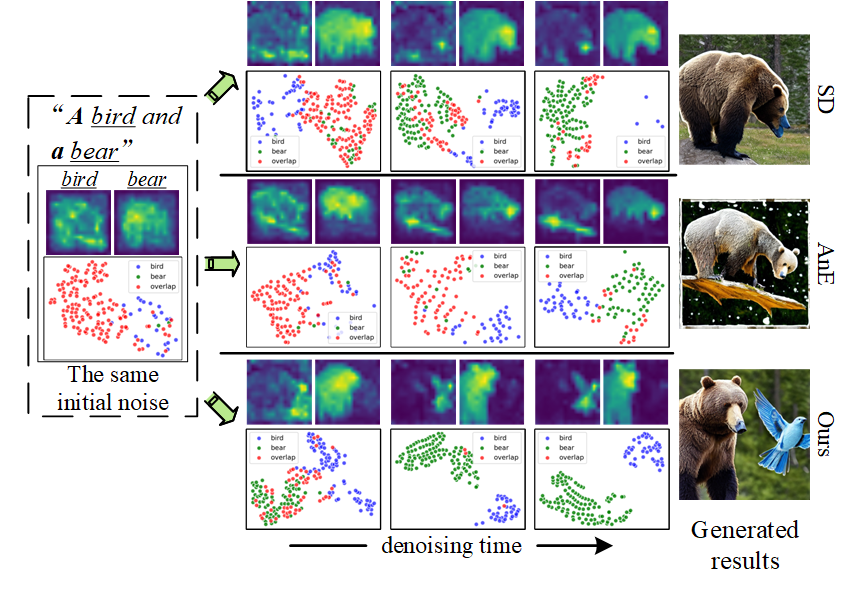}
	\caption{Visual analysis on how the isolation loss of AI-T2I addresses inter-token overlap. AI-T2I achieves desirable subject generation compared to existing works, \emph{i.e.}, Stable Diffusion \cite{sd} and AnE \cite{attend}.}
	\label{fig:max_cluster_compare}
\end{figure}	
	
\noindent \textbf{Visual Analysis on Isolation Loss and Aggregation Loss.} To shed more light into the isolation loss of AI-T2I, we perform additional visual analysis with the same initial noise via T-SNE, along with generated images and cross-attention maps during the denoising process. Fig. \ref{fig:max_cluster_compare} illustrates that, for AI-T2I, the activations belonging to different subjects (\emph{i.e.}, bird and bear) gradually move away from each other as the denoising time goes, yielding desirable subject generation; while the compared approaches (Stable Diffusion \cite{sd} and AnE \cite{attend}) still suffer from numerous overlapping point (red point). Such fact further confirms the \emph{benefits} of the isolation loss equipped with Euclidean distance to mitigate inter-token overlap issue.

	To validate the effectiveness of ${\cal L}_{agg-sub}$ for aggregation loss, we conduct additional ablation study with the same initial noise via the following cases: \textbf{A}: removing ${\cal L}_{agg-sub}$ and ${\cal L}_{agg-attr}$, \textbf{B}: removing ${\cal L}_{agg-attr}$, which reports the generated images and corresponding cross-attention maps during the denoising process. Fig. \ref{fig:aggregate} illustrates that case \textbf{A} bears the scattering issue for identical subjects, \emph{i.e.}, the extra black suitcase, compared to case \textbf{B}, implying that, with ${\cal L}_{agg-sub}$, our AI-T2I is capable of achieving noticeable subject aggregation.

\noindent \textbf{Ablation Study on $r$ and $n$.} Tab. \ref{tb:ablation_r_n} shows ablation study on $r$ and $n$ in aggregation loss. The number of grouping regions $n$ has a greater impact on performance than their radius $r$. Additionally, different datasets perform best with different $r$, likely due to inherent size differences. Animals exhibit substantial variations in body size and limb motion; hence, the Animals-Animals subdataset utilizes a time-varying parameter $r$. In contrast, as man-made objects generally possess highly regular shapes, a fixed $r$ is optimal for the Objects-Objects subdataset.

\begin{figure}[t]
		\centering
		\includegraphics[width=0.8\linewidth]{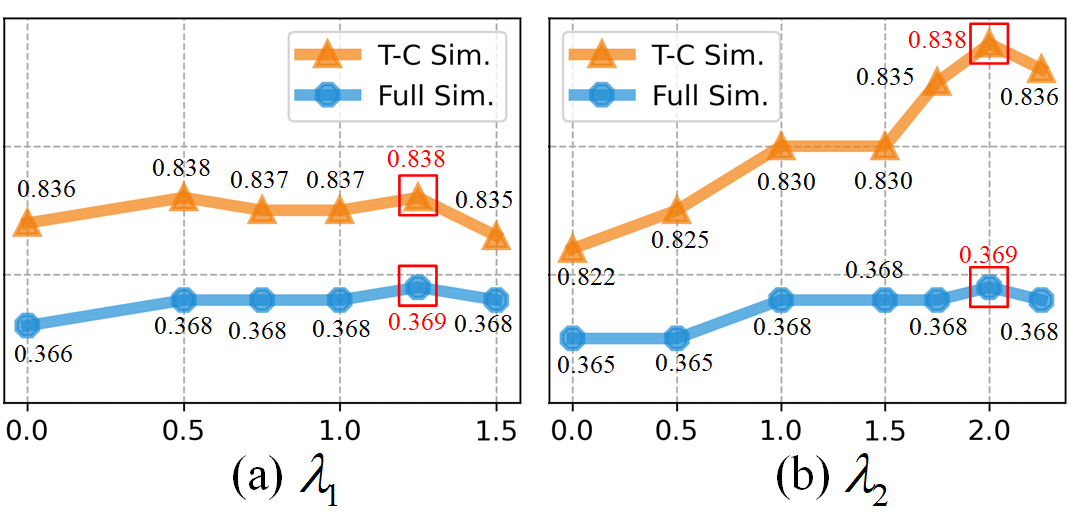}
		\caption{Ablation studies on the parameters (a) $\lambda_1$ and (b) $\lambda_2$ for the aggregation and isolation losses via Full Sim. and T-C Sim.
		}
		\label{fig:ablation_study_weight}
	\end{figure}

\noindent \textbf{How does aggregation loss collaborate with isolation loss?} The parameters $\lambda_1$ and $\lambda_2$ in Eq. \ref{eq:overall_loss} are utilized to balance the aggregation and isolation loss, to address inter-subject-token overlap and intra-subject-token scattering. We further validate the effectiveness by setting varied $\lambda_1=\{0,0.5,0.75,1,1.25,1.5\}$ with fixed $\lambda_2=2$, apart from $\lambda_2=\{0,0.5,1,1.5,1.75,2,2.25\}$ with fixed $\lambda_1=1.25$. Fig. \ref{fig:ablation_study_weight} illustrates that the optimal performance is achieved with $\lambda_1^*=1.25$ and $\lambda_2^*=2$, implying that carefully adjusting $\lambda_1$ and $\lambda_2$ can yield extra gains. Such fact further confirms the \emph{benefits} of close relationship between aggregation and isolation losses.

	\section{Conclusion}
	In this paper, we propose a novel Aggregating-and-Isolating cross-attention approach to diffusion models for Text-to-Image synthesis, dubbed AI-T2I, which improves text-to-image alignment by addressing intra-token scattering and inter-token overlap issues in cross-attention maps via proposed aggregation and isolation losses. Extensive experiments demonstrate that our AI-T2I outperforms existing methods across multiple benchmarks and generalizes well to multiple downstream tasks. 
	
	\begin{table}[t]
		\small
		\centering
		\setlength{\abovecaptionskip}{-0.1cm}
		\caption{Ablation study on $r$ and $n$ in aggregation loss. $r=[a, b]$ denotes that $r$ increases from an initial value of $a$ to a maximum of $b$ over the denoising timesteps. The best results are reported with \textbf{boldface}.} 
		\begin{tabular}{@{}l c c c|c c c@{}} 
			\toprule
			
			\rowcolor{gray!30}
			\multicolumn{7}{@{}l}{\textbf{Animals-Animals}} \\
			
			\multirow{2}{*}{T-C Sim.} & r=[1, 8] & r=[2, 8] & r=[3, 8] & n=2 & n=3 & n=4  \\
			

			~ & 0.832	&\textbf{0.834}	&0.831 &\textbf{0.834} 	&0.829 & 0.817\\
			
			\midrule 
			\rowcolor{gray!30}
			\multicolumn{7}{@{}l}{\textbf{Objects-Objects}} \\
			\multirow{2}{*}{T-C Sim.}& r=4 & r=5 & r=6 & n=2 & n=3 & n=4  \\
			~ & 0.827	&\textbf{0.828}	&0.825 &0.823 	&\textbf{0.828} & 0.827\\

			\bottomrule 
		\end{tabular}
		\label{tb:ablation_r_n}
		\vspace{-2mm}
	\end{table}

\begin{IEEEbiography}[{\includegraphics[width=1in,height=1.25in,clip,keepaspectratio]{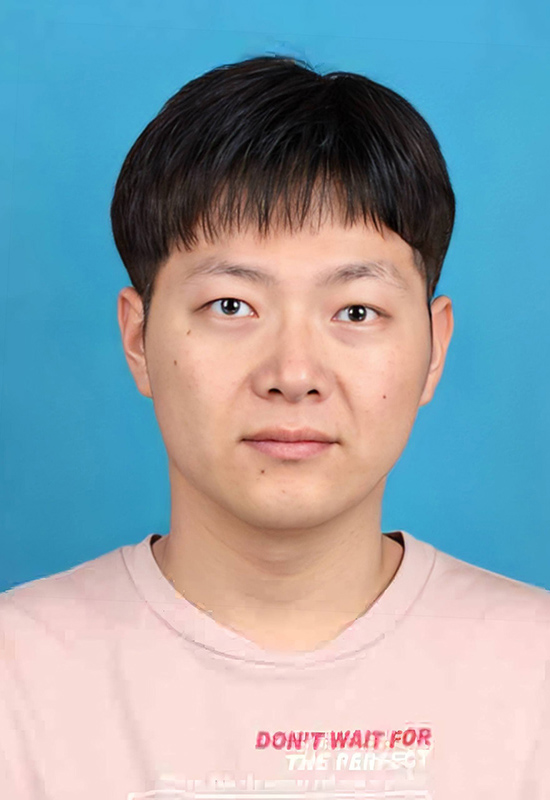}}]{Shipeng Cao}
	is currently working toward the PhD degree with the Hefei University of Technology, Hefei,
	China. His research interests include computer vision and deep learning.
\end{IEEEbiography}

\begin{IEEEbiography}[{\includegraphics[width=1in,height=1.25in,clip,keepaspectratio]{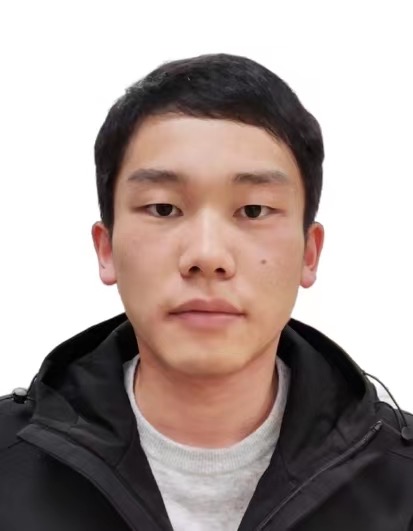}}]{Biao Qian}
is currently a Post Ph.D at Tsinghua University. He received the Ph.D degrees from the Hefei University of Technology, Hefei, China. His primary research interests include computer vision, model compression, knowledge distillation and lightweight multi-modal large language models. He has published several research papers including IEEE Transactions on Pattern Analysis and Machine Intelligence, CVPR, ECCV, AAAI, and IEEE Transactions on Image Processing.
\end{IEEEbiography}

\begin{IEEEbiography}[{\includegraphics[width=1in,height=1.25in,clip,keepaspectratio]{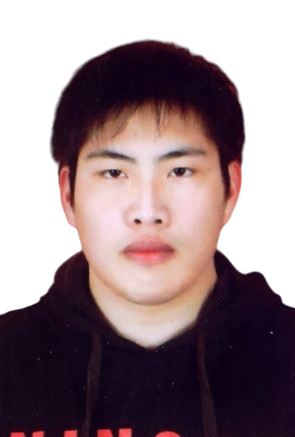}}]{Haipeng Liu}
	received the PhD degree with the Hefei University of Technology, Hefei,
	China. His research interests include computer vision and deep learning. He has published several research papers including CVPR, ACM MM, and NeurIPS.
\end{IEEEbiography}

\begin{IEEEbiography}[{\includegraphics[width=1in,height=1.25in,clip,keepaspectratio]{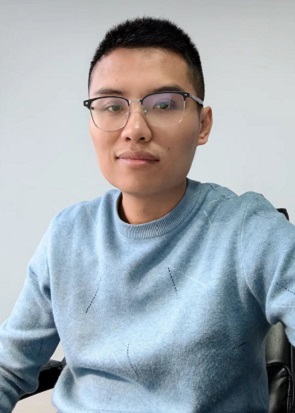}}]{Yang Wang} (Senior Member, IEEE) is currently a full professor with the Hefei University of Technology, China. He has published more than 100 research papers, including IEEE Transactions on Pattern Analysis and Machine Intelligence, IJCV, Artificial Intelligence (Elsevier), ICML, NeurIPS, CVPR, ICCV, ECCV, KDD, SIGIR, AAAI, IJCAI, ACM Multimedia, IEEE Transactions on	Image Processing, ACM TKDD,  ACM Transactions on Information Systems, Machine Learning (Springer), and IEEE Transactions on Knowledge and Data Engineering. He recieved the hot paper award for Science China Information Sciences, together with 10 papers selected as	ESI highly cited papers (Top 1\%). His research has gained more than 8000	Google Scholar citations and H-index 45. He is also an associate editor of ACM Transactions on Information Systems.
\end{IEEEbiography}

\begin{IEEEbiography}[{\includegraphics[width=1in,height=1.25in,clip,keepaspectratio]{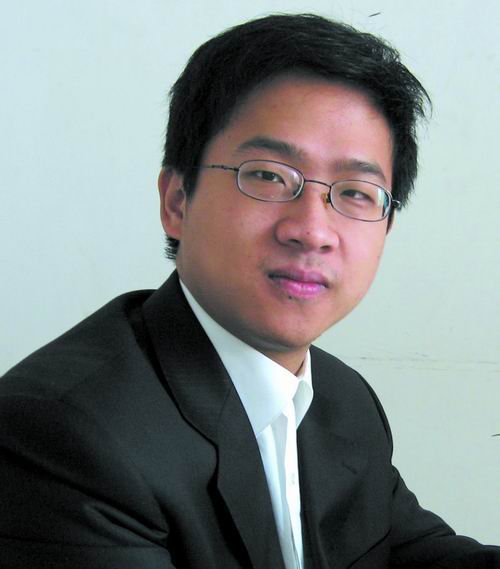}}]{Meng Wang}	(Fellow, IEEE) received the BE and PhD degrees from the Special Class for the Gifted Young, Department of Electronic Engineering and Information Science, University of Science and Technology of China (USTC), Hefei, China, in 2003 and 2008, respectively. He is currently a professor with the Hefei University of Technology, China. He received paper prizes or awards from ACM MM 2009 (the Best Paper Award), ACM MM 2010 (the Best Paper Award), ACM MM 2012 (the Best Demo Award), ICDM 2014 (the Best Student Paper Award), SIGIR 2015 (the Best Paper Honorable Mention), IEEE TMM 2015 and 2016 (the Prize Paper Award	Honorable Mention), IEEE SMC 2017 (the Best Transactions Paper Award), and ACM TOMM 2018 (the Nicolas D. Georganas Best Paper Award). He is currently an associate editor of IEEE Transactions on Pattern Analysis and Machine Intelligence, IEEE Transactions on Knowledge and Data Engineering, IEEE Transactions on Multimedia, and IEEE Transactions on Neural Networks and Learning Systems.
\end{IEEEbiography}


\begin{thebibliography}{1}
	
	\bibitem{song2020denoising}
	J.~Song, C.~Meng, and S.~Ermon, ``Denoising diffusion implicit models,''
	\emph{arXiv preprint arXiv:2010.02502}, 2020.
	
	\bibitem{dhariwal2021diffusion}
	P.~Dhariwal and A.~Nichol, ``Diffusion models beat gans on image synthesis,''
	\emph{Advances in neural information processing systems}, vol.~34, pp.
	8780--8794, 2021.
	
	
	\bibitem{dong2023prompt}
	W.~Dong, S.~Xue, X.~Duan, and S.~Han, ``Prompt tuning inversion for text-driven
	image editing using diffusion models,'' in \emph{Proceedings of the IEEE/CVF
		International Conference on Computer Vision}, 2023, pp. 7430--7440.
	
	\bibitem{mao2023guided}
	J.~Mao, X.~Wang, and K.~Aizawa, ``Guided image synthesis via initial image
	editing in diffusion model,'' in \emph{Proceedings of the 31st ACM
		International Conference on Multimedia}, 2023, pp. 5321--5329.
	
	\bibitem{guo2024initno}
	X.~Guo, J.~Liu, M.~Cui, J.~Li, H.~Yang, and D.~Huang, ``Initno: Boosting
	text-to-image diffusion models via initial noise optimization,'' in
	\emph{Proceedings of the IEEE/CVF Conference on Computer Vision and Pattern
		Recognition}, 2024, pp. 9380--9389.
	
	
	\bibitem{attend}
	H.~Chefer, Y.~Alaluf, Y.~Vinker, L.~Wolf, and D.~Cohen-Or, ``Attend-and-excite:
	Attention-based semantic guidance for text-to-image diffusion models,''
	\emph{ACM Transactions on Graphics (TOG)}, vol.~42, no.~4, pp. 1--10, 2023.
	
	\bibitem{astar}
	A.~Agarwal, S.~Karanam, K.~Joseph, A.~Saxena, K.~Goswami, and B.~V. Srinivasan,
	``A-star: Test-time attention segregation and retention for text-to-image
	synthesis,'' in \emph{Proceedings of the IEEE/CVF International Conference on
		Computer Vision}, 2023, pp. 2283--2293.
	
	\bibitem{prompt2prompt}
	A.~Hertz, R.~Mokady, J.~Tenenbaum, K.~Aberman, Y.~Pritch, and D.~Cohen-or,
	``Prompt-to-prompt image editing with cross-attention control,'' in \emph{The
		Eleventh International Conference on Learning Representations}.
	
	\bibitem{conform}
	T.~H.~S. Meral, E.~Simsar, F.~Tombari, and P.~Yanardag, ``Conform: Contrast is
	all you need for high-fidelity text-to-image diffusion models,'' in
	\emph{Proceedings of the IEEE/CVF Conference on Computer Vision and Pattern
		Recognition}, 2024, pp. 9005--9014.
	
	\bibitem{predicated}
	K.~Sueyoshi and T.~Matsubara, ``Predicated diffusion: Predicate logic-based
	attention guidance for text-to-image diffusion models,'' in \emph{Proceedings
		of the IEEE/CVF Conference on Computer Vision and Pattern Recognition}, 2024,
	pp. 8651--8660.
	
	\bibitem{linguistic}
	R.~Rassin, E.~Hirsch, D.~Glickman, S.~Ravfogel, Y.~Goldberg, and G.~Chechik,
	``Linguistic binding in diffusion models: Enhancing attribute correspondence
	through attention map alignment,'' \emph{Advances in Neural Information
		Processing Systems}, vol.~36, pp. 3536--3559, 2023.
	
	\bibitem{structure}
	W.~Feng, X.~He, T.-J. Fu, V.~Jampani, A.~R. Akula, P.~Narayana, S.~Basu, X.~E.
	Wang, and W.~Y. Wang, ``Training-free structured diffusion guidance for
	compositional text-to-image synthesis,'' in \emph{The Eleventh International
		Conference on Learning Representations}, 2023.
	

	\bibitem{huang2017real}
	H.~Huang, H.~Wang, W.~Luo, L.~Ma, W.~Jiang, X.~Zhu, Z.~Li, and W.~Liu,
	``Real-time neural style transfer for videos,'' in \emph{Proceedings of the
		IEEE Conference on Computer Vision and Pattern Recognition}, 2017, pp.
	783--791.
	
	\bibitem{ding2021cogview}
	M.~Ding, Z.~Yang, W.~Hong, W.~Zheng, C.~Zhou, D.~Yin, J.~Lin, X.~Zou, Z.~Shao,
	H.~Yang \emph{et~al.}, ``Cogview: Mastering text-to-image generation via
	transformers,'' \emph{Advances in neural information processing systems},
	vol.~34, pp. 19\,822--19\,835, 2021.
	
	\bibitem{ramesh2021zero}
	A.~Ramesh, M.~Pavlov, G.~Goh, S.~Gray, C.~Voss, A.~Radford, M.~Chen, and
	I.~Sutskever, ``Zero-shot text-to-image generation,'' in \emph{International
		conference on machine learning}.\hskip 1em plus 0.5em minus 0.4em\relax Pmlr,
	2021, pp. 8821--8831.
	
	\bibitem{freecontrol}
	S.~Mo, F.~Mu, K.~H. Lin, Y.~Liu, B.~Guan, Y.~Li, and B.~Zhou, ``Freecontrol:
	Training-free spatial control of any text-to-image diffusion model with any
	condition,'' in \emph{Proceedings of the IEEE/CVF Conference on Computer
		Vision and Pattern Recognition}, 2024, pp. 7465--7475.
	
	\bibitem{towards}
	B.~Liu, C.~Wang, T.~Cao, K.~Jia, and J.~Huang, ``Towards understanding cross
	and self-attention in stable diffusion for text-guided image editing,'' in
	\emph{Proceedings of the IEEE/CVF Conference on Computer Vision and Pattern
		Recognition}, 2024, pp. 7817--7826.
	
	\bibitem{sd}
	R.~Rombach, A.~Blattmann, D.~Lorenz, P.~Esser, and B.~Ommer, ``High-resolution
	image synthesis with latent diffusion models,'' in \emph{Proceedings of the
		IEEE/CVF conference on computer vision and pattern recognition}, 2022, pp.
	10\,684--10\,695.
	
	\bibitem{zhang2025enhancing}
	Y.~Zhang, T.~T. Tzun, L.~W. Hern, and K.~Kawaguchi, ``Enhancing semantic
	fidelity in text-to-image synthesis: Attention regulation in diffusion
	models,'' in \emph{European Conference on Computer Vision}.\hskip 1em plus
	0.5em minus 0.4em\relax Springer, 2025, pp. 70--86.
	
	\bibitem{xie2023boxdiff}
	J.~Xie, Y.~Li, Y.~Huang, H.~Liu, W.~Zhang, Y.~Zheng, and M.~Z. Shou, ``Boxdiff:
	Text-to-image synthesis with training-free box-constrained diffusion,'' in
	\emph{Proceedings of the IEEE/CVF International Conference on Computer
		Vision}, 2023, pp. 7452--7461.
	
	\bibitem{phung2024grounded}
	Q.~Phung, S.~Ge, and J.-B. Huang, ``Grounded text-to-image synthesis with
	attention refocusing,'' in \emph{Proceedings of the IEEE/CVF Conference on
		Computer Vision and Pattern Recognition}, 2024, pp. 7932--7942.
	
	\bibitem{park2024energy}
	G.~Y. Park, J.~Kim, B.~Kim, S.~W. Lee, and J.~C. Ye, ``Energy-based cross
	attention for bayesian context update in text-to-image diffusion models,''
	\emph{Advances in Neural Information Processing Systems}, vol.~36, 2024.
	
	\bibitem{zhang2024object}
	Y.~Zhang, P.~Yu, and Y.~N. Wu, ``Object-conditioned energy-based attention map
	alignment in text-to-image diffusion models,'' in \emph{European Conference
		on Computer Vision}.\hskip 1em plus 0.5em minus 0.4em\relax Springer, 2024,
	pp. 55--71.
	
	\bibitem{dahary2024yourself}
	O.~Dahary, O.~Patashnik, K.~Aberman, and D.~Cohen-Or, ``Be yourself: Bounded
	attention for multi-subject text-to-image generation,'' in \emph{European
		Conference on Computer Vision}.\hskip 1em plus 0.5em minus 0.4em\relax
	Springer, 2024, pp. 432--448.
	
	\bibitem{binyamin2024make}
	L.~Binyamin, Y.~Tewel, H.~Segev, E.~Hirsch, R.~Rassin, and G.~Chechik, ``Make
	it count: Text-to-image generation with an accurate number of objects,''
	\emph{arXiv preprint arXiv:2406.10210}, 2024.
	
	\bibitem{li2023divide}
	Y.~Li, M.~Keuper, D.~Zhang, and A.~Khoreva, ``Divide \& bind your attention for
	improved generative semantic nursing,'' \emph{arXiv preprint
		arXiv:2307.10864}, 2023.
	

	
	\bibitem{orgad2023editing}
	H.~Orgad, B.~Kawar, and Y.~Belinkov, ``Editing implicit assumptions in
	text-to-image diffusion models,'' in \emph{Proceedings of the IEEE/CVF
		International Conference on Computer Vision}, 2023, pp. 7053--7061.
	
	
	\bibitem{huang2024smartedit}
	Y.~Huang, L.~Xie, X.~Wang, Z.~Yuan, X.~Cun, Y.~Ge, J.~Zhou, C.~Dong, R.~Huang,
	R.~Zhang \emph{et~al.}, ``Smartedit: Exploring complex instruction-based
	image editing with multimodal large language models,'' in \emph{Proceedings
		of the IEEE/CVF Conference on Computer Vision and Pattern Recognition}, 2024,
	pp. 8362--8371.
	
	\bibitem{nam2024dreammatcher}
	J.~Nam, H.~Kim, D.~Lee, S.~Jin, S.~Kim, and S.~Chang, ``Dreammatcher:
	appearance matching self-attention for semantically-consistent text-to-image
	personalization,'' in \emph{Proceedings of the IEEE/CVF Conference on
		Computer Vision and Pattern Recognition}, 2024, pp. 8100--8110.
	
	\bibitem{guo2024smooth}
	J.~Guo, X.~Xu, Y.~Pu, Z.~Ni, C.~Wang, M.~Vasu, S.~Song, G.~Huang, and H.~Shi,
	``Smooth diffusion: Crafting smooth latent spaces in diffusion models,'' in
	\emph{Proceedings of the IEEE/CVF Conference on Computer Vision and Pattern
		Recognition}, 2024, pp. 7548--7558.
	
	\bibitem{liu2024towards}
	B.~Liu, C.~Wang, T.~Cao, K.~Jia, and J.~Huang, ``Towards understanding cross
	and self-attention in stable diffusion for text-guided image editing,'' in
	\emph{Proceedings of the IEEE/CVF conference on computer vision and pattern
		recognition}, 2024, pp. 7817--7826.
	
	\bibitem{clip}
	A.~Radford, J.~W. Kim, C.~Hallacy, A.~Ramesh, G.~Goh, S.~Agarwal, G.~Sastry,
	A.~Askell, P.~Mishkin, J.~Clark \emph{et~al.}, ``Learning transferable visual
	models from natural language supervision,'' in \emph{International conference
		on machine learning}.\hskip 1em plus 0.5em minus 0.4em\relax PmLR, 2021, pp.
	8748--8763.
	
	\bibitem{li2022blip}
	J.~Li, D.~Li, C.~Xiong, and S.~Hoi, ``Blip: Bootstrapping language-image
	pre-training for unified vision-language understanding and generation,'' in
	\emph{International conference on machine learning}.\hskip 1em plus 0.5em
	minus 0.4em\relax PMLR, 2022, pp. 12\,888--12\,900.
	
	\bibitem{2025selfcross}
	Q.~Weimin, W.~Jieke, and T.~Meng, ``Self-cross diffusion guidance for
	text-to-image synthesis of similar subjects,'' in \emph{CVPR}, 2025.
	
	\bibitem{t2ibench}
	K.~Huang, C.~Duan, K.~Sun, E.~Xie, Z.~Li, and X.~Liu, ``T2i-compbench++: An
	enhanced and comprehensive benchmark for compositional text-to-image
	generation,'' \emph{IEEE Transactions on Pattern Analysis and Machine
		Intelligence}, 2025.
	
	\bibitem{hao2024conceptexpress}
	S.~Hao, K.~Han, Z.~Lv, S.~Zhao, and K.-Y.~K. Wong, ``Conceptexpress: Harnessing
	diffusion models for single-image unsupervised concept extraction,'' in
	\emph{European Conference on Computer Vision}.\hskip 1em plus 0.5em minus
	0.4em\relax Springer, 2024, pp. 215--233.
	
	\bibitem{li2023gligen}
	Y.~Li, H.~Liu, Q.~Wu, F.~Mu, J.~Yang, J.~Gao, C.~Li, and Y.~J. Lee, ``Gligen:
	Open-set grounded text-to-image generation,'' in \emph{Proceedings of the
		IEEE/CVF conference on computer vision and pattern recognition}, 2023, pp.
	22\,511--22\,521.
	
	\bibitem{2018COCO}
	H.~Caesar, J.~Uijlings, and V.~Ferrari, ``Coco-stuff: Thing and stuff classes
	in context,'' in \emph{2018 IEEE/CVF Conference on Computer Vision and
		Pattern Recognition (CVPR)}, 2018.
	
	\bibitem{2020YOLOv4}
	A.~Bochkovskiy, C.~Y. Wang, and H.~Y.~M. Liao, ``Yolov4: Optimal speed and
	accuracy of object detection,'' 2020.
	
	\bibitem{Liu_2024_CVPR}
	H.~Liu, Y.~Wang, B.~Qian, M.~Wang, and Y.~Rui, ``Structure matters: Tackling
	the semantic discrepancy in diffusion models for image inpainting,'' in
	\emph{Proceedings of the IEEE/CVF Conference on Computer Vision and Pattern
		Recognition (CVPR)}, June 2024, pp. 8038--8047.
	
	\bibitem{Qian_2023_CVPR}
	B.~Qian, Y.~Wang, R.~Hong, and M.~Wang, ``Adaptive data-free quantization,'' in
	\emph{Proceedings of the IEEE/CVF Conference on Computer Vision and Pattern
		Recognition (CVPR)}, June 2023, pp. 7960--7968.
	
	\bibitem{qian2023rethinking}
	B.~Qian, Y.~Wang, R.~Hong, and M.~Wang, ``Rethinking data-free quantization as a zero-sum game,'' in
	\emph{Proceedings of the AAAI conference on artificial intelligence},
	vol.~37, no.~8, 2023, pp. 9489--9497.
	
	\bibitem{10476709}
	Y.~Wang, B.~Qian, H.~Liu, Y.~Rui, and M.~Wang, ``Unpacking the gap box against
	data-free knowledge distillation,'' \emph{IEEE Transactions on Pattern
		Analysis and Machine Intelligence}, vol.~46, no.~9, pp. 6280--6291, 2024.
	
	\bibitem{One_Stone}
	H. Liu, Y. Wang, and M. Wang,
	``One Stone with Two Birds: A Null-Text-Null Frequency-Aware Diffusion Models for Text-Guided Image Inpainting,'' \emph{Advances in Neural Information
		Processing Systems}, 2025.

\bibitem{wu2018and}
L.~Wu, Y.~Wang, J.~Gao, and X.~Li, ``Where-and-when to look: Deep siamese
attention networks for video-based person re-identification,'' \emph{IEEE
	Transactions on Multimedia}, vol.~21, no.~6, pp. 1412--1424, 2018.

\bibitem{xu2024sgdm}
Y.~Xu, X.~Xu, H.~Gao, and F.~Xiao, ``Sgdm: an adaptive style-guided diffusion
model for personalized text to image generation,'' \emph{IEEE Transactions on
	Multimedia}, vol.~26, pp. 9804--9813, 2024.
	

\bibitem{hou2025clip}
Y.~Hou, W.~Zhang, Z.~Zhu, and H.~Yu, ``Clip-gan: Stacking clips and gan for
efficient and controllable text-to-image synthesis,'' \emph{IEEE Transactions
	on Multimedia}, 2025.

\bibitem{cheng2022vision}
Q.~Cheng, K.~Wen, and X.~Gu, ``Vision-language matching for text-to-image
synthesis via generative adversarial networks,'' \emph{IEEE Transactions on
	Multimedia}, vol.~25, pp. 7062--7075, 2022.

\bibitem{yang2024dmf}
B.~Yang, X.~Xiang, W.~Kong, J.~Zhang, and Y.~Peng, ``Dmf-gan: Deep multimodal
fusion generative adversarial networks for text-to-image synthesis,''
\emph{IEEE Transactions on Multimedia}, vol.~26, pp. 6956--6967, 2024.

\bibitem{xu2024sgdm}
Y.~Xu, X.~Xu, H.~Gao, and F.~Xiao, ``Sgdm: an adaptive style-guided diffusion
model for personalized text to image generation,'' \emph{IEEE Transactions on
	Multimedia}, vol.~26, pp. 9804--9813, 2024.

\bibitem{zheng2025semantic}
J.~Zheng, N.~Xu, W.~Li, J.~Jiang, and X.~Zhang, ``Semantic-spatial attention
for refined object placement in text-to-image synthesis,'' \emph{IEEE
	Transactions on Multimedia}, 2025.

\bibitem{taghipour2025box}
A.~Taghipour, M.~Ghahremani, M.~Bennamoun, A.~M. Rekavandi, H.~Laga, and
F.~Boussaid, ``Box it to bind it: Unified layout control and attribute
binding in text-to-image diffusion models,'' \emph{IEEE Transactions on
	Multimedia}, 2025.

\bibitem{zhang2025aligning}
Z.~Zhang, S.~Zhang, L.~Shen, Y.~Zhan, Y.~Luo, H.~Hu, B.~Du, Y.~Wen, and D.~Tao,
``Aligning text-to-image diffusion models with constrained reinforcement
learning,'' \emph{IEEE Transactions on Pattern Analysis and Machine
	Intelligence}, 2025.

\bibitem{flux}
Black Forest Labs. Official weights of FLUX.1 dev. https : / / huggingface . co / black - forest labs/FLUX.1-dev, 2024. Accessed: 2024-11-14.

\bibitem{hunyuandit2024}
Z.~Li, J.~Zhang, Q.~Lin, J.~Xiong, Y.~Long, X.~Deng, Y.~Zhang, X.~Liu,
M.~Huang, Z.~Xiao \emph{et~al.}, ``Hunyuan-dit: A powerful multi-resolution
diffusion transformer with fine-grained chinese understanding,'' \emph{arXiv
	preprint arXiv:2405.08748}, 2024.

\bibitem{qwenimage2025}
C.~Wu, J.~Li, J.~Zhou, J.~Lin, K.~Gao, K.~Yan, S.-m. Yin, S.~Bai, X.~Xu,
Y.~Chen \emph{et~al.}, ``Qwen-image technical report,'' \emph{arXiv preprint
	arXiv:2508.02324}, 2025.	
	
\bibitem{liu2025few}
H.~Liu, G.~Li, M.~Gao, X.~Zhen, F.~Zheng, and Y.~Wang, ``Few-shot referring
video single-and multi-object segmentation via cross-modal affinity with
instance sequence matching,'' \emph{International Journal of Computer
	Vision}, vol. 133, no.~8, pp. 5610--5628, 2025.

\bibitem{liu2026chipdiff}
H.~Liu, Z.~Song, Y.~Wang, B.~Hu, and Y.~Wang, ``Chipdiff: Staged diffusion
model with loss gradient guidance for chinese ink painting style transfer,''
\emph{Pattern Recognition}, p. 113309, 2026.

\bibitem{qian2022switchable}
B.~Qian, Y.~Wang, H.~Yin, R.~Hong, and M.~Wang, ``Switchable online knowledge
distillation,'' in \emph{European Conference on Computer Vision}.\hskip 1em
plus 0.5em minus 0.4em\relax Springer, 2022, pp. 449--466.

\bibitem{zhang2025survey}
X.~Zhang, X.~Wei, W.~Hu, J.~Wu, J.~Wu, W.~Zhang, Z.~Zhang, Z.~Lei, and Q.~Li,
``A survey on personalized content synthesis with diffusion models,''
\emph{Machine Intelligence Research}, vol.~22, no.~5, pp. 817--848, 2025.

\bibitem{jiang2025cmsl}
Y.~Jiang, Y.~Lyu, B.~Peng, W.~Wang, and J.~Dong, ``Cmsl: Cross-modal style
learning for few-shot image generation,'' \emph{Machine Intelligence
	Research}, vol.~22, no.~4, pp. 752--768, 2025.	


\bibitem{wang2022progressive}
Y.~Wang, J.~Peng, H.~Wang, and M.~Wang, ``Progressive learning with multi-scale
attention network for cross-domain vehicle re-identification,'' \emph{Science
	China Information Sciences}, vol.~65, no.~6, p. 160103, 2022.	
	
\end{thebibliography}
\end{document}